\documentclass[acmsmall]{acmart}

\AtBeginDocument{%
  \providecommand\BibTeX{{%
    \normalfont B\kern-0.5em{\scshape i\kern-0.25em b}\kern-0.8em\TeX}}}

\setcopyright{acmcopyright}
\copyrightyear{2018}
\acmYear{2018}
\acmDOI{10.1145/1122445.1122456}

\usepackage{multirow}
\usepackage{bm}
\usepackage{xspace}
\usepackage{balance}
\usepackage{amsmath}
\usepackage{graphicx}
\usepackage{url}

\newcommand{\eat}[1]{}
\newcommand{\paratitle}[1]{\vspace{1ex}\noindent \textbf{#1}}
\let\oldhat\hat
\renewcommand{\vec}[1]{\bm{#1}}
\renewcommand{\hat}[1]{\oldhat{\mathbf{#1}}}
\renewcommand{\matrix}[1]{\bm{#1}}
\newcommand{\eg}{\emph{e.g., }\xspace}
\newcommand{\rf}{\emph{rf. }\xspace}

\newcommand{\ie}{\emph{i.e., }\xspace}

\newcommand{\etal}{\emph{et al. }\xspace}
\newcommand{\aka}{\emph{a.k.a., }\xspace}
\newcommand\BibTeX{B\textsc{ib}\TeX}

\acmJournal{TOIS}
\acmVolume{x}
\acmNumber{x}
\acmArticle{x}
\acmYear{2020}
\acmMonth{0x}

\begin{document}

\title{Target Guided  Emotion Aware Chat Machine}

\author{Wei Wei$^\dagger$~}
\affiliation{Cognitive Computing and Intelligent Information Processing (CCIIP) Laboratory, School of Computer Science and Technology, Huazhong University of Science and Technology.}
\author{Jiayi Liu$^\dagger$~}
\affiliation{Cognitive Computing and Intelligent Information Processing (CCIIP) Laboratory, School of Computer Science and Technology, Huazhong University of Science and Technology}
\author{Xianling Mao~}
\affiliation{School of Computer Science and Technology, Beijing Institute of Technology}
\author{Guibing Guo~}
\affiliation{Software College, Northeastern University}
\author{Feida Zhu~}
\affiliation{School of Information Systems, Singapore Management University}
\author{Pan Zhou~}
\affiliation{School of Computer Science and Technology, Huazhong University of Science and Technology}
\author{Yuchong Hu~}
\affiliation{ School of Computer Science and Technology, Huazhong University of Science and Technology}
\author{Shanshan Feng~}
\affiliation{Inception Institute of Artificial Intelligence Abu Dhabi, UAE}
\thanks{An earlier version of this article was presented at CIKM 2019, titled: ``Emotion-aware Chat Machine: Automatic Emotional Response Generation for Human-like Emotional Interaction'' \cite{wei2019emotion}.}

{\let\thefootnote\relax\footnotetext{$\dagger$Corresponding author: weiw@hust.edu.cn;\quad liujiayi7@hust.edu.cn.}}

\renewcommand{\shortauthors}{Wei, et al.}

\begin{abstract}
The consistency of a response to a given post at \textbf{\emph{semantic}}-level and \textbf{\emph{emotional}}-level is essential for a dialogue system to deliver human-like interactions. However, this challenge is not well addressed in the literature, since most of the approaches neglect the emotional information conveyed by a post while generating responses. This article addresses this problem and proposes a \emph{unified} end-to-end neural architecture, which is capable of simultaneously encoding the \emph{semantics} and the \emph{emotions} in a post, and leveraging target information to generate more intelligent
responses with appropriately expressed emotions. Extensive experiments on real-world data demonstrate that the proposed method outperforms the state-of-the-art methods in terms of both content coherence and emotion appropriateness.
\end{abstract}


\maketitle

\section{Introduction}\label{sec:intro}
Dialogue systems in practice are typically built for various applications like emotional interaction~\cite{mayer2008human}, customer service~\cite{zhou2020design} and question answering~\cite{wei2011integrating,wei2016exploring}, which can be roughly categorized into three classes, \ie chit-chat chatbot, task-oriented chatbot and domain-specific chatbot. For example, a task-specific chatbot can serve as a customer consultant, while a chit-chat chatbot is commonly designed for convincingly simulating how a human would respond as a conversational partner. In fact, most recent work on response generation in chitchat domain can be summarized as follows, \ie retrieval-based, matching-based, or statistical machine learning based approaches~\cite{wallace2003elements,wilcox2011beyond,ritter2011data,ji2014information}. Recently, with the increasing popularity of deep learning, many research efforts have been dedicated to employing an encode-decoder architecture~\ie Sequence-to-sequence (\emph{Seq2seq}) models~\cite{sutskever2014sequence,cho2014learning}, to map a post to the corresponding response with little hand-crafted features or domain-specific knowledge for the conversation generation problem~\cite{shang2015neural,vinyals2015neural}. Subsequently, several variants of \emph{Seq2seq} models are also proposed to address different issues~\cite{serban2016building,xing2017topic,qian2017assigning,mou2016sequence}.

Despite the great progress made in neural dialogue generation, a general fact is that several works have been reported to automatically incorporate emotional factors for dialogue systems. In fact, several empirical studies have proven that chatbots with the ability of emotional communication with humans are essential for  enhancing user satisfaction~\cite{martinovski2003breakdown,prendinger2005using,callejas2011predicting}. To this end,  it is highly valuable and desirable to develop an emotion-aware chatbot that is capable of perceiving/expressing emotions and emotionally interacting with the interlocutors. In literature, Zhou \etal \cite{zhou2018emotional} successfully build an emotional chat machine (ECM) that is capable of generating emotional responses according to a pre-defined emotion category, and several similar efforts are also made by \cite{peng2019topic,huang2018automatic}. Besides, \cite{zhou2017mojitalk} proposed by Zhou~\etal utilizes emojis to control the emotional response generation process within conditional variational auto-encoder (CAVE) framework.

\begin{figure}[!t]
\centering
\includegraphics[scale=0.6]{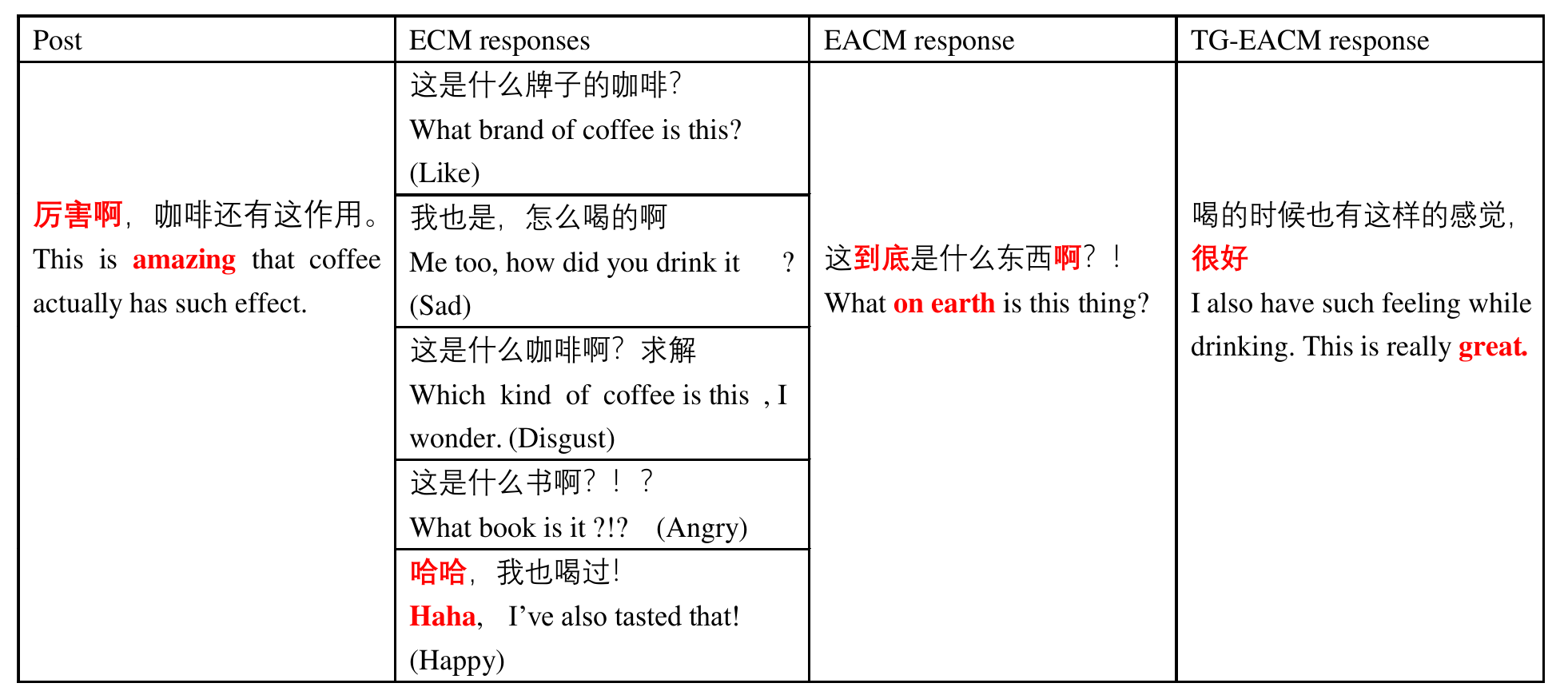}
\caption{Conversation examples of ECM, EACM and TG-EACM.
ECM is capable of generating multiple responses via giving different predefined emotions, which is called semi-automatic response generation in this paper. In comparison, EACM and TG-EACM are called automatic generation models. Additionally, from Fig.1 we can observe that, the response generated by TG-EACM is more comfortable than the one generated by EACM, owing to TG-EACM chooses an optimal emotion ({\it Like}), rather than ({\it Other}), which is selected by EACM.}
\label{fig1}
\vspace{-0.65cm}
\end{figure}

Nevertheless, these mentioned approaches cannot work well owing to the following facts:
(i) These approaches solely generate the emotional responses based on a pre-specified label (or emoji) as shown in Figure~\ref{fig1}, which is unrealistic in practice as the well-designed dialogue systems need to wait for a manually selected emotion category for response generation;
(ii) The generation process, which is apparently divided into two parts, would significantly hurt the smoothness and quality of generating responses;
and (iii) As shown in Figure~\ref{fig1},  some emotionally-inappropriate responses (even conflicts) might apparently affect the interlocutor's satisfaction. Thereby, here we argue that fully exploiting the information of the given post and leveraging the guidance from the target to supervise the generating process is definitely beneficial for automatically generating responses with the optimal emotion~(\rf ``TG-EACM response'' generated by our method shown in Figure~\ref{fig1}).

Previous methods greatly contribute to emotion-expressing conversation generation problem, however, they are insufficient and several issues emerge when trying to fully-address this problem.
\textbf{\emph{First}}, it is not easy to model human emotion from a given sentence due to the semantic sparsity in natural language. Psychological studies~\cite{plutchik1980general,plutchik2001nature} demonstrate that human emotion is quite complex and one sentence may contain multi-types of emotions with different intensities. For example, A traveler might comment a hotel with ``The environment is not bad, but the location is too partial''. Another example is, a man may congratulate his friend on getting a new job, ``Congratulations, but you're gonna leave to another city for living, what's a pity!'', both of which conveys complex emotions (\eg dual emotions), such as positive (\eg happy) and negative (\eg regretful) emotions. As such, solely using the post's emotion label is insufficient for smoothly generating emotion-aware response, and thus we consider to learn more helpful guidance information from the target. 
\textbf{\emph{Second}}, it is difficult for a model to decide the optimal response emotion for generation, as the emotions between the post and the response are not 1-1 mapping, and thus it may be not reasonable to directly make the emotion label of the post to be as the one of the response. Actually, the emotion selection process is not only by the post's emotion but also by its semantic meaning.
\textbf{\emph{Third}}, it is also problematic to design a unified model that can generate plausible emotional sentence without sacrificing grammatical fluency and semantic coherence~\cite{zhou2018emotional}. Hence, the response generation problem faces a significant challenge: that is, how to effectively leverage handy information from the post as well as the target to automatically learn the emotion interaction pattern for emotion-aware response generation within a unified model.

In this paper, we propose a innovative \underline{T}arget-\underline{G}uided \underline{E}motion-\underline{A}ware \underline{C}hat \underline{M}achine (named \textsf{TG-EACM} for short), which is capable of perceiving other interlocutor's feeling (\ie post’s emotion) and generating plausible response with the optimal emotion category (\ie response's emotion).Specifically, TG-EACM is based on a unified \emph{Seq2seq} architecture with a self-attention enhanced emotion selector and an emotion-biased response generator, to simultaneously modeling the post's emotional and semantic information for automatically generating appropriate response. Targets (\ie responses in the training data) are used as guidance to boost the learning ability of the emotion interaction pattern. Experiments on the public datasets demonstrate the effectiveness of our proposed method, in terms of two different types of evaluation metrics, \ie \emph{automatic metric}, which is used to measure the diversity of words in the generated sentences, and \emph{human evaluation}, which is used to decide whether the generated responses' emotions are appropriate according to human annotations. The main contributions of this research are summarized as follows:

\begin{enumerate}
    \item It advances the current emotion conversation generation problem from a new perspective, namely emotion-aware response generation, by taking account of the emotional interactions between interlocutors.
    \item It also proposes an innovative generation model (\ie TG-EACM) that is capable of extracting post's emotional and semantic information and leveraging guidance from the targets in order to generate intelligible responses with appropriate emotions.
            In specific,  we use a prior network and a posterior network (which is not used in the original model) for mitigating the gap between the training part and the testing one, as the posterior network can learn a better intermedia encoder both from the post and the given response, which can effectively guide the model to learn the emotion transition  between the post and response, and then the prior network can help more accurately modeling the post’s semantic and emotion information during testing, and in the ablation study, we shows that the proposed method can significantly improve the evaluation performance in terms of automatic metric and human evaluation.
    \item It conducts extensive experiments on a real-word dataset, which demonstrate the proposed approach outperforms the state-of-the-art methods at both \emph{semantic}-level and \emph{emotional}-level.
\end{enumerate}

\section{Related Works}
\label{sec:r_work}
\paratitle{General model}.
The early work on conversational models is initially proposed by Joseph in 1960s, called ELIZA~\cite{Weizenbaum:1966}, which primarily builds a dialogue system based on the handcrafted templates and heuristic rules. However, the manual process is subjective, time-consuming and tedious, and thus making it difficult for large-scale conversation generation problem. Recently, with the exponential growth of available human-to-human conversation data, many data-driven approaches are proposed for the
conversation generation task, which generally falls into two categories, \ie \emph{retrieval}-based and \emph{generation}-based.

\textbf{\emph{Retrieval}}-based methods aim at choosing an appropriate response from the historical data via measuring the matching degree between the post and the candidate response. There already exist several efforts dedicated to research on \emph{single-turn} response selection where the input is a single post~\cite{wang2013dataset,hu2014convolutional,lu2013deep}. Later, many deep learning based approaches are proposed for \emph{multi-turn} response selection problem
via calculating context-response matching score, such as dual LSTM model~\cite{lowe2015ubuntu}, deep learning to respond architecture~\cite{yan2016learning}, sequential matching network (SMN)~\cite{yan2016sequential}, multi-view matching model~\cite{zhou2016multiview}, deep attention matching network (DAM)~\cite{zhou2018multi} and multi-representation fusion network (MRFN)~\cite{Tao2019mrfn}.

With advances in deep learning, \textbf{\emph{generation}}-based models have become the most promising solution to many conversation generation tasks, through regarding the conversation generation as a \emph{monolingual translation} task that encodes a post into a fixed-length vector from which a decoder generates a response, such as \emph{chit-chat} or \emph{domain-specific} conversation~\cite{zhang2018tailored}. The mainstream of representative methods are usually based on Sequence-to-sequence (\emph{Seq2seq}) architecture with attention mechanism~\cite{shang2015neural,sordoni2015neural,vinyals2015neural}, as its strong ability in effectively addressing the sequence mapping issues \cite{sutskever2014sequence,cho2014learning,bahdanau2014neural}. Several variants of \emph{Seq2seq} model are also proposed, such as hierarchical recurrent model~\cite{serban2016building,serban2017hierarchical} or topic-aware model~\cite{xing2017topic}.

Besides, there also exist several lines of studies on different aspects of conversation generation problem, such as: (i) \emph{Context-consistency}, which aims to endow the chatbots with particular personality traits for addressing the context-consistency problem, \eg persona-based model~\cite{li2016persona} and identity-coherent model~\cite{qian2017assigning}; (ii) \emph{Diversity}, the goal of which is to enhance the diversity and informativeness of generated responses, \eg Maximum Mutual Information (MMI) based model~\cite{li2015diversity} or enhanced beam-search based model~\cite{li2016simple,vijayakumar2016diverse}; and (iii) \emph{Commonsense Knowledge}, which takes account of static graph attention to incorporate commonsense knowledge for chatbots~\cite{zhou2018commonsense}. In addition, Zhang~\etal~\cite{zhang2018tailored} also propose different solutions for two classical conversation scenarios, \ie chit-chat and domain-specific conversation. However, all of these methods do not consider \emph{emotion} factor for \emph{emotion}-aware conversation generation as we do in this work.

\paratitle{Emotion-aware model}.
The methods of emotion-aware conversation generation aims at perceiving other interlocutor’s feeling (\ie post’s emotion) and generating a plausible response with the optimal emotion category (\ie response’s emotion) \cite{li2019empgan}. Ghosh~\etal~\cite{ghosh2017affect} propose \emph{affect language model} to generate texts conditioned on a specified affect categories with controllable affect strength. Hu~\etal~\cite{hu2017toward} present a combination of Variational Auto-Encoder (VAE) and holistic attribute discriminators to generate sentences with certain types of sentiment and tense, and which is however mainly built for emotional text generation task.
However, they are not designed for emotion-aware conversation generation problem.

In recent years, several proposals~\cite{zhou2018emotional,zhou2017mojitalk,huang2018automatic} approach the problem of generating more intelligent responses with appropriately expressed emotions towards a given post, which is of great significance for successfully building human-like dialogue systems. Zhou~\etal~\cite{zhou2018emotional} develop an Emotional Chat Machine (ECM) model composed of three different mechanisms (\ie \emph{emotion embedding}, \emph{internal memory} and \emph{external memory}) for response generation, according to a designated emotion category. Similarly, Zhou~\etal~\cite{zhou2017mojitalk} propose a reinforcement learning based approach with Conditional Variational Auto-Encoder (CVAE) architecture to generate responses conditioned on several given emojis. In ~\cite{huang2018automatic}, Huang~\etal consider emotion factors for response generation via different emotion injecting methods. And more recently, in \cite{song2019generating} Song~\etal proposes to leverage lexicon-based attention mechanism to encourage expressing emotion words. However, all of such models are designed based on a \emph{semi}-automatic architecture that needs to pre-define an optimal response emotion category for generation. Besides, they also ignore the post's emotion information during generation process, which easily leads to an insufficient and inhumane response. In contrast, our proposed model is able to automatically learn the emotion interaction mode within a unified framework for more human-like (emotion-aware) response generation, through effectively and explicitly making use of the emotional and semantic information from the given post.

\paratitle{Others}.
Indeed, there also exist numerous attempts to extend the basic encoder-decoder architecture for improving the performance of \emph{Seq2seq} model. For example,
Bahdanau \etal~\cite{bahdanau2014neural} extend a Bi-directional Long Short-Term Memory (Bi-LSTM) network with attention mechanism for long-sentence generation,
which is able to automatically predict the target word by (soft-)searching for relevant parts in the context, rather than explicitly modeling such parts as a hard segment. Luong~\etal \cite{luong2015effective} thoroughly evaluate the effectiveness of different types of attention mechanisms, \ie global/local attention with different alignment functions. Furthermore, several studies employ the techniques from machine translation domain for improving performance, \eg self-attention mechanism~\cite{lin2017structured,vaswani2017attention}, which has been proven that can yield large gains in terms of  BLEU~\cite{papineni2002bleu}, as compared to the state-of-the-art methods. Jean~\etal~\cite{jean2014using} utilize large vocabularies and back-off dictionaries to address the out-of-vocabulary (OOV) problem, and achieve the encouraging results by using a sampled soft-max method for tackling the increasing complexity during decoding (caused by large-scale vocabularies).
Though these works have successfully build a solid foundation for future studies based on the optimized \emph{Seq2seq} model, the purposes of such approaches are different from our current work.

\section{Proposed Model}
\label{sec:model}
\setlength{\tabcolsep}{5pt}
\begin{table}[t]
\centering\caption{Notations and Definitions.}
\footnotesize\label{tbl:symbol and definition}{
\vspace{0ex}
\begin{tabular}{cp{170pt}} \hline\hline
\textbf{Notation} & \textbf{Definition}   \\ \hline
        \textbf{Input} & \\ \hline
        $\vec{x}$ & Post utterance, $\vec{x}= \{x_1,x_2,\cdots, x_{T}\}$; \\
        $\vec{y}$ & Response utterance, $\vec{y} = \{y_1,y_2,\cdots, y_{T^{'}}\};$ \\
        $\vec{e_p}$ & Post's emotion label denoted by a multi-hot vector, \eg $(0,1,1,0,0)$;\\
        $\vec{e_r}$ & Response's emotion label denoted by a multi-hot vector, \eg $(1,1,0,0,0)$; \\ \hline
        $\vec{e_r^{*}}$ & The selected response's emotion over emotion categories, denoted by a multi-hot vector, \eg $(0.1,0.3,0.5,0,0.1)$; \\ \hline
        \textbf{Latent Variable} & \\ \hline
        $\vec{h}_p^t,\vec{h}_e^t,\vec{h}_r^t$ & Hidden states of prior encoder, intermediate encoder and recognition encoder at $t$-th step; \\
        $\widetilde{\vec{h}}_{pe},\widetilde{\vec{h}}_{re}$ & Fused representation of prior network and recognition network; \\
        $\oldhat{\vec{e_p}},\oldhat{\vec{e_r}},\oldhat{\vec{e_{r'}}}$ & Estimated post's emotion vector based on \emph{prior} encoder; and response's emotion vectors from the \emph{prior} network and the \emph{recognition} network, respectively;  \\
        $\vec{c}_t$ & Context vector for the generator calculated by attention mechanism at step $t$; \\
        $\vec{s}_t$ & Hidden states of the generator at step $t$; \\
        $\vec{V}_e$ & Embedded response emotion vector; \\  \hline
        \textbf{Loss Function} & \\ \hline
        {$\mathcal{L}_p, \mathcal{L}_{r}, \mathcal{L}_{r'}$} & Cross entropy loss over post's emotion label; and response emotion label from the \emph{prior} network and the \emph{recognition} network, respectively; \\ 
        $\mathcal{L}_{KL}$ &  Kullback-Leibler divergence loss between the prior network and the recognition network; \\ 
        $\mathcal{L}_{NLL}$ & Negative log likelihood loss for generated response. \\ \hline\hline
\end{tabular}}
\vspace{0ex}
\end{table}

\subsection{Preliminary: Seq2Seq with Attention Model}
Recently, generation-based methods have received a considerable amount of attention in dialogue generation domain, \eg Seq2seq-attention model~\cite{bahdanau2014neural},
as proven in many studies~\cite{bahdanau2014neural,cho2014learning,sutskever2014sequence}, which can promote the quality of generated sentences via dynamically attending on the key information of the input post during decoding. As Seq2seq-attention architecture is a widely-adopted solution for response generation problem, therefore our approach is also mainly based on it to generate responses. Next, we firstly illustrate this basic model in principle.

The seq2seq-attention model is typically a deep RNN-based architecture with an encoder and a decoder. The encoder takes the given post sequence  $\vec{x}= \{x_1,x_2,\cdots, x_{T}\}$~($T$ is the length of the post) as inputs, and maps them into a series of hidden representations $\vec{h} = (\vec{h}_1,\vec{h}_2,\cdots, \vec{h}_{T})$.  The decoder then decodes them to generate a word sequence, \ie $\vec{y} = \{y_1, y_2, \cdots, y_{T^{'}}\}$, where $T^{'}$ is the length of the output sequence, and it may differ from $T$.

In more detail, the context representation $\vec{c}_{t}$ of the post sequence $\vec{x}$ is computed by parameterizing the encoder hidden vector $\vec{h}$ with different attention scores~\cite{bahdanau2014neural}, that is,
\begin{equation}
  \vec{c}_{t}=\sum_{j=1}^{T}\vec{\alpha}(\vec{s}_{t-1},\vec{h_j})\cdot \vec{h_j},
\end{equation}
where $\vec{\alpha}(.,.)$ is a weighted coefficient estimated by each encoder token's contribution  to the target word $y_t$.
The decoder iteratively updates its state $\vec{s}_t$ using previously generated word $\vec{y}_{t-1}$, namely,
\begin{equation}
\label{eq:over_opt}
  \vec{s}_t = f(\vec{s}_{t-1},\vec{y}_{t-1},\vec{c_t}),\quad\quad t  = 1, 2, \cdots, {T^{'}},
\end{equation}
where $f(.,.,.)$ is a non-linear transformation of RNN cells~(\eg LSTM~\shortcite{hochreiter1997long} or GRU~\shortcite{cho2014learning}).

Then, the probability of generating  the $t$-th token $y_t$ conditioned on the input sequence $\vec{x}$ and the previous predicted word sequence $y_{1:t-1}$
is computed by
\begin{equation}
\begin{array}{ll}
\Pr \left( y_t|y_{1:t-1}, \vec{x}\right) =g(\vec{y}_{t-1},\vec{s_t},\vec{c_t}),
\end{array}
\end{equation}
where $g(.,.,)$ is a function (\eg $\mbox{softmax}$) to produce valid probability distribution for sampling the next word of the output sequence.

\subsection{Problem Definition and Overview}
\label{subsec:pro_def_overview}
\begin{figure*}[!t]
\centering
\includegraphics[scale=0.25]{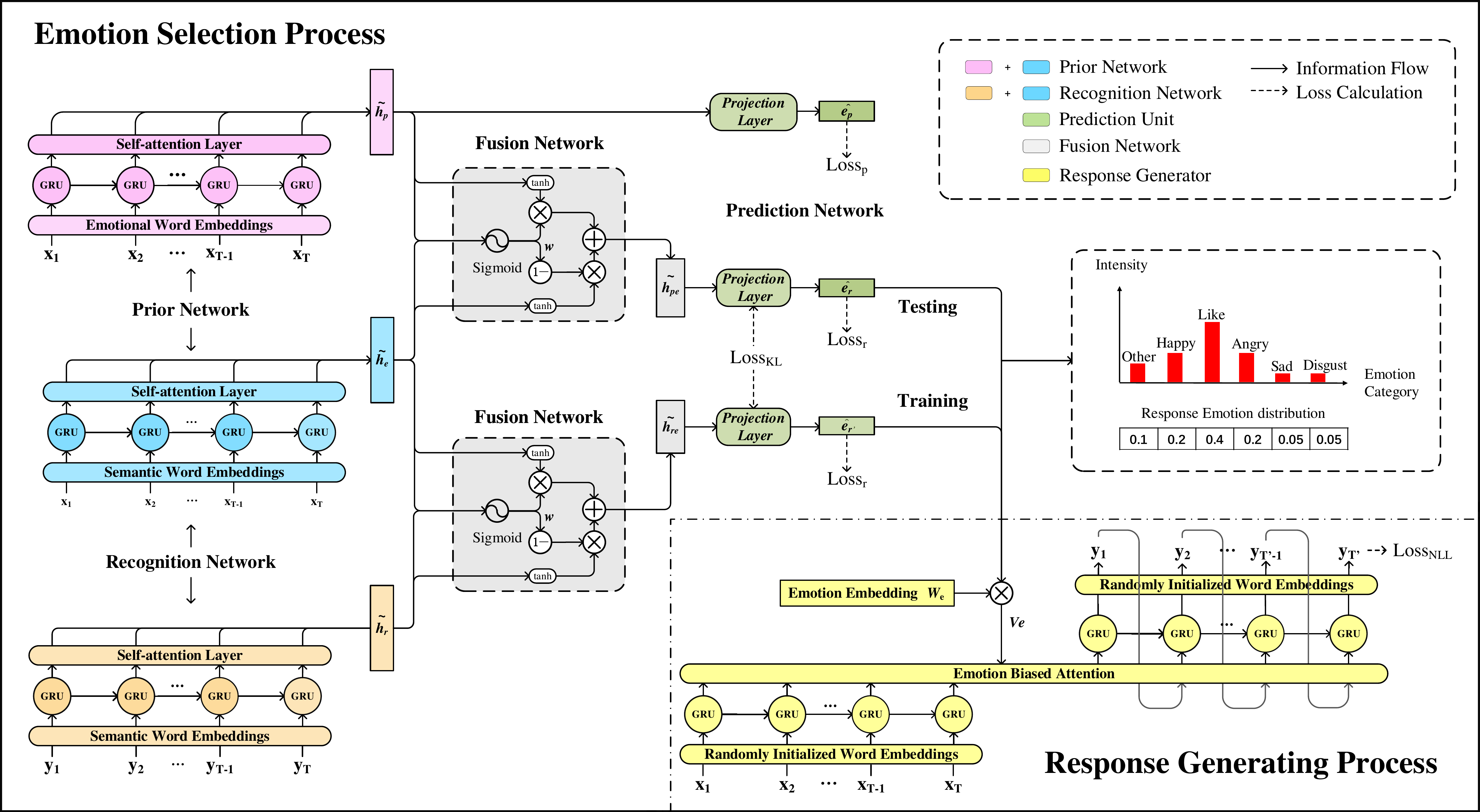}
\caption{Framework of TG-EACM.}
\label{fig2}

\end{figure*}

In this paper, our task is to perceive the emotion involved in the input post and incorporate it into the generation process for \emph{automatically} producing both \emph{semantically reasonable} and \emph{emotionally appropriate} response. Hence, our conversation generation problem is formulated as follows: given a post $\vec{x}\!=\!\{x_1,\cdots, x_{T}\}$ and an emotion category $\vec{e_p}$ for a responder, the problem of automatic emotion-aware response generation aims to generate a  corresponding response sequence $\vec{y} = \{y_1,y_2,\cdots, y_{T^{'}}\}$ with proper emotion $\vec{e_r}$.
To tackle the problem, we first need to answer two key questions as follows. For ease of reference, some notations and their definitions are listed in Table~\ref{tbl:symbol and definition}.
\begin{itemize}
  \item \textbf{1.}~How to automatically
select an appropriate response emotion for generation according to the emotion and semantic information derived from the given post, which can be defined as
\begin{equation}
\label{eq-best-emotion}
\vec{e_r}^{*}\leftarrow  \mathop{\arg\max}_{\vec{e_r}\in \mathcal{E}_{c}} \Pr(\vec{e_r}|\vec{x},\vec{e_p}),
\end{equation}
where $\mathcal{E}_{c}$ is the $K$-dimensional vector space of the emotions ($K$ represents the number of emotions);
and $\vec{e_r}^{*}$ is the generated emotion distribution over emotion categories~(\rf Fig.~\ref{fig2}).
\item \textbf{2.}~How to generate an emotion-aware response via the obtained emotion $\vec{e_r}^{*}$ and the input post $\vec{x}$, namely, the problem is how to estimate the probability of the response sequence $\vec{y}$ with a standard LSTM-LM formulation given the encoded post sequence $\vec{x}$ with the selected emotion $\vec{e_r}^{*}$,
\begin{equation}
\label{entire-eq-generate}
\begin{array}{ll}
\hspace{-10pt}\Pr(\vec{y}|\vec{x},\vec{e_r}^{*})
&\hspace{-6pt}=\Pr(y_1,y_2,\cdots,y_{T^{'}}|\vec{x},\vec{e_p})\\
&\hspace{-6pt}=\prod^{T^{'}}_{t=1}{\Pr(y_t|y_{1:t-1},\vec{x},\vec{e_r}^{*})}
\end{array}
\end{equation}
Therefore, the problem is transformed to estimate the probability of predicting the target word $y_t$ by given the previous word sequence $y_{1:t-1}$, the given post $\vec{x}$ and the optimal response emotion $\vec{e_r}^{*}$, \ie
\begin{equation}
\vec{y_t}^{*}\leftarrow \mathop{\arg\max} \Pr(y_{t}|y_{1:t-1},\vec{x},\vec{e_r}^{*}).
\end{equation}
\end{itemize}

To address the above two problems, we propose a novel approach named \underline{T}arget-\underline{G}uided \underline{E}motion-\underline{A}ware \underline{C}hat \underline{M}achine (\ie \textsf{TG-EACM}) within an unified framework presented in Fig.~\ref{fig2}, which primarily consists of two subcomponents, \ie \emph{emotion selector} and \emph{response generator}. More concretely, the \emph{emotion selector} is in charge of the emotion selection process, \ie yielding an emotion distribution over $\mathcal{E}_{c}$, which is use for \emph{response generator} to generate a corresponding response based on the input post. We will detail them in the following sections, respectively.

\paratitle{Remark}.
Actually, previous methods usually assume the emotion of the generated response is derived from an unique emotion category\footnote{Here we follow the work~\cite{zhou2018emotional} to use ESTC dataset with the emotion category (\ie \{Angry, Disgust, Happy, Like, Sad, Other\}) for experimental comparison.}
and is thus denoted by a one-hot vector, where the corresponding emotion is labeled with $1$, and $0$ for others. However, we observe that emotions of responses are more complicated in practice, that is, there are no simple emotion mappings between the post and the response~(\ie not $1$-$1$ mapping). As such, we use a emotion probability distribution over $\mathcal{E}_{c}$ to model the emotion of a response, rather than directly mapping it into a single emotion category. Note that we follow academical conventions to distinguish the word ``sentiment'' and ``emotion'', and in our paper ``sentiment'' is defined as the effect of ``emotion'', and thus such as ``happy’, ``anger'' and ``love'', which are defined as emotions and the sentiments like ``positive'', ``negative'' and ``neutral'' are defined as sentiments.

\subsection{Target Guided Emotion Selector}
\subsubsection{Overview}
Given an input post $\vec{x}$ with its emotion label~$\vec{e_p}$, one way to infer the response emotion is to approximately estimate the conditional probability over a emotional distribution by using a simple RNN-based architecture. But as mentioned in Section~\ref{subsec:pro_def_overview}, there are no simple emotion mappings between the post and the response~(\ie not one-to-one mapping).

To address the problem, we propose a novel approach named \emph{\underline{T}arget-\underline{G}uided \underline{E}motion \underline{S}elector} (\textsf{TG-ES})
to model the emotional interaction by approximately estimating the posterior probability conditioned on the post and  its response within a \emph{prior-recognition} architecture~(as shown in Fig.~\ref{fig2}), which consists of three sub-components, \ie \emph{prior network}, \emph{recognition network} and \emph{fusion network}\footnote{Note that we label the different components with different color for better distinction in Fig.~\ref{fig2}.}. Hereinafter, Eq.~\eqref{eq-best-emotion} can be transformed to approximately estimate the response emotion probability distribution conditioned on post $\vec{x}$ and response $\vec{y}$, namely,
\begin{equation}
\label{eq-best-emotion-new}
\vec{e_r}^{*}\leftarrow  \mathop{\arg\max}_{\vec{e_r}\in \mathcal{E}_{c}} \Pr(\vec{e_r}|\vec{x},\vec{y}).
\end{equation}

Specifically, here  the \emph{recognition} network is used to calculate the \emph{posterior} distribution conditioned on the post and the corresponding response,
to incorporate the target (\ie \emph{response}) information as a guidance for \emph{prior network} for more accurately modeling the emotional interaction. Then, the \emph{fusion} network is used to balance the contributions derived from different types of information (\eg the semantic and the emotional information of the post), and finally a prediction network is employed to generate the optimal emotion vector over the fusion information. Additionally, it is worth noting that the discrepancy between the \emph{prior} distribution and the \emph{posterior} distribution makes the model hard to train, and we thus utilize \emph{Kullback-Leibler}~(KL) loss to minimize the difference between the \emph{prior} network and the \emph{recognition} network, which will be detailed in the following section (\rf Section~\ref{sec:fusion_prediction_network}).


\subsubsection{Self-Attention Based Encoder}

\begin{figure}[!t]
\centering
\includegraphics[scale=0.5]{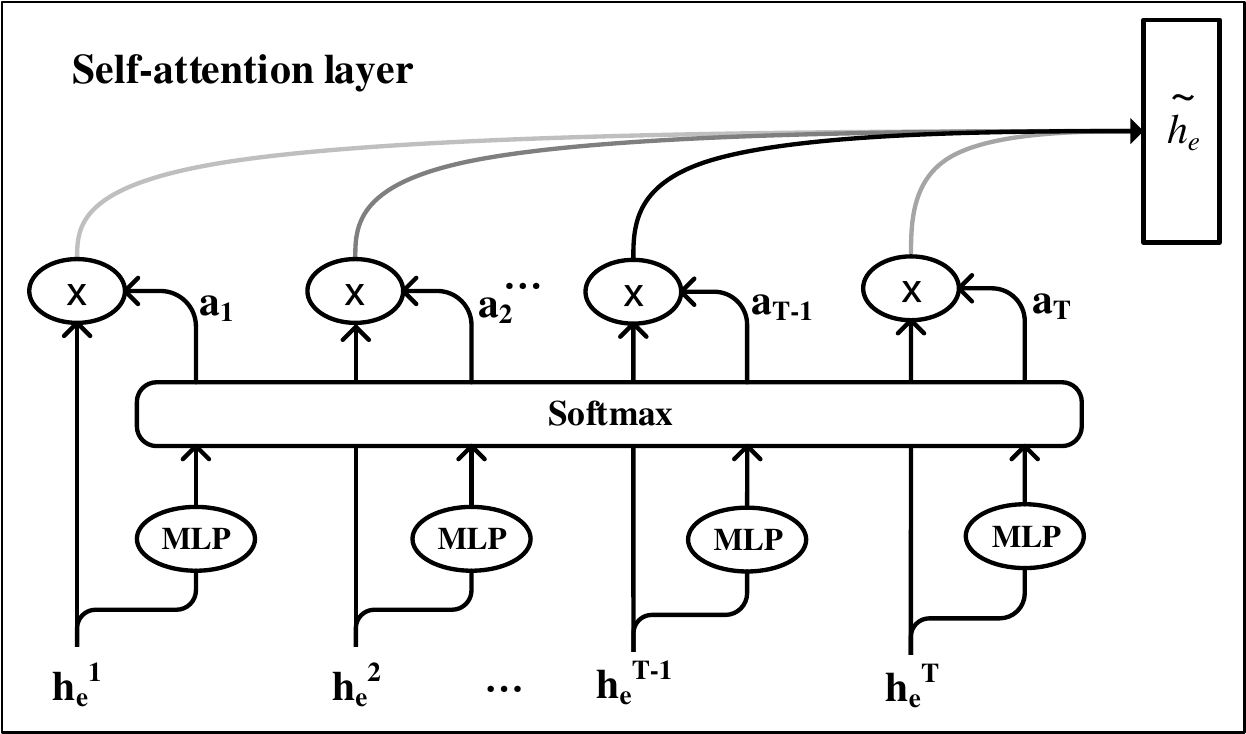}
\caption{\textbf{Example}.~An illustration of self-attention layer for \emph{intermediate} encoder.}
\label{fig3}
\vspace{0ex}
\end{figure}

To model the emotion interaction during selection process, we propose a novel \emph{ self-attention} based encoder to explicitly extract the semantic and emotional information of the post and the response for representation. In fact, there are three different types of encoders in our model, \ie \emph{prior} encoder, \emph{recognition} encoder and \emph{intermediate} encoder, which are used to encode post's emotional and semantic information, as well as the response's emotional information, respectively. It is worth nothing that the \emph{intermediate} encoder is simultaneously used in the \emph{prior} network and the \emph{recognition} network, as it is used to make the emotion distribution of the posterior network and the prior network consistent.

Specifically, these encoders are firstly implemented through \emph{GRU} cells for extracting auxiliary information from post sequence $\vec{x}= (x_1,x_2,\cdots, x_{T})$
and response sequence $\vec{y} = \{y_1,y_2,\cdots, y_{T^{'}}\}$, and map them into hidden representations using the following formulas:
\begin{align}
  \vec{h}_p^t & = \mathbf{GRU}(\vec{h}_p^{t-1},x_t), \\
  \vec{h}_e^t & = \mathbf{GRU}(\vec{h}_e^{t-1},x_t), \\
  \vec{h}_r^t & = \mathbf{GRU}(\vec{h}_r^{t-1},y_t),
\end{align}
where $\vec{h}_p^t$, $\vec{h}_e^t$ and $\vec{h}_r^t$ denote the hidden states of the \emph{prior} encoder, the \emph{intermediate} encoder and the \emph{recognition} encoder at $t$-th time step, respectively.

To enhance the representation power of the hidden states, we introduce \emph{self-attention} mechanism~\cite{lin2017structured} to enable such encoders to attend to \emph{information-rich} words in the utterances, and then obtain the attended hidden states $\widetilde{\vec{h}}_p$, $\widetilde{\vec{h}}_e$ and $\widetilde{\vec{h}}_r$ by calculating:
\begin{align}
\widetilde{\vec{h}}_p & = \sum_{i=1}^{T}a^p_i\vec{h}_p^i,\\
\widetilde{\vec{h}}_e & = \sum_{i=1}^{T}a^e_i\vec{h}_e^i,  \\
\widetilde{\vec{h}}_r & = \sum_{i=1}^{T}a^r_i\vec{h}_r^i,
\end{align}
where $a^p_i$, $a^e_i$, $a^r_i$ are the weights of the $i$-th hidden states of different encoders (\ie $\vec{h}_p$, $\vec{h}_e$, $\vec{h}_r$), which are calculated by feeding $\vec{h}_p^i$, $\vec{h}_e^i$ and $\vec{h}_r^i$ into a multi-layer perceptron with a \emph{softmax} layer to ensure that the sum of all the weights equals to $1$,
\begin{align}
a^p_i & = \textbf{softmax}(\vec{V_p}\tanh(\vec{W_p}(\vec{h}_p^i)^\top), \\
a^e_i & = \textbf{softmax}(\vec{V_e}\tanh(\vec{W_e}(\vec{h}_e^i)^\top), \\
a^r_i & = \textbf{softmax}(\vec{V_r}\tanh(\vec{W_r}(\vec{h}_r^i)^\top).
\end{align}

Differentiate from the \emph{prior} encoder and the \emph{recognition} encoder, the \emph{intermediate} encoder aims to ensure the post's \emph{prior} emotional distribution is consistent with the post's \emph{posterior} emotional distribution. As such, let $\mathcal{L}_p$ be a loss function that imposes a cross entropy loss on the top of the emotion hidden state $\widetilde{\vec{h}}_p$, namely, passing the emotion hidden state through a linear layer and a sigmoid layer to project it into an emotion distribution over $\mathcal{E}_{c}$, and the cross entropy loss is calculated as follows,
\begin{align}
\oldhat{\vec{e}_p} = \sigma(\vec{W_p}\widetilde{\vec{h}}_p + b ),\\
\mathcal{L}_p = - \vec{e}_p {\rm log}(\oldhat{\vec{e}_p}),
\end{align}
where $\vec{e}_p$ is a \emph{multi}-hot representation of the post's emotion vector (\eg $(0,1,1,1,0)$) since it may contain multiple emotions.


\subsubsection{Emotional and Semantic Word Embedding}
To force the emotion selector to focus on different aspects of the post's auxiliary information, we consider to use the \emph{recognition} encoder and the \emph{intermediate} encoder for semantic embedding, as well as the prior encoder for emotional embedding, respectively.

Most of existing traditional word embedding approaches (\eg \emph{word2vec}~\cite{mikolov2013efficient}) usually only model the syntactic context of words for making the words with similar syntactic context close, which may result in mapping words with totally \emph{opposite} sentiment polarity to neighboring word vectors, \eg \emph{good} and \emph{bad}. As such, here we consider to employ {\it SSWE}~\cite{tang2014learning} for emotional embedding, which has been proven that is effective in learning \emph{sentiment}-specific word embedding by incorporating the supervision from sentiment polarity of text to encode sentiment information in the continuous representation of words, which is capable of mapping words with similar sentiment polarity to neighboring word vectors, such \emph{good} and \emph{beautiful}. As such, here we consider to make use of {\it SSWE} and {\it word2vec} for emotional embedding and semantic embedding in our model, respectively.

Based on such embedding settings, the \emph{prior} network (\ie \emph{prior} encoder and \emph{intermediate} encoder) can learn to extract from both the post's \emph{emotional} and \emph{semantic} information, and the \emph{recognition} network (\ie \emph{recognition} encoder and \emph{intermediate} encoder) is able to offer the semantic guidance from both the post and the response, and the prior-recognition network can work interactively to effectively supervise the modeling of the emotion interaction during selection process.

\subsubsection{Fusion and Prediction Network}
\label{sec:fusion_prediction_network}

Simply mapping $\oldhat{\vec{e}_p}$ to the response emotion category $\vec{e_r}$ is insufficient to model the emotion interaction process between partners, as we cannot choose emotion category only based on post's emotion category under some circumstances. In fact, some posts expressing negative feelings like sad are inappropriate to be replied with the same emotion, such as ``It's a pity you can't come with us'' or ``I'm so sad that you broke my heart''. Therefore, we not only consider the post's emotional information, but also take into account its semantic meaning by combining the hidden states from the prior encoder and intermediate encoder (\ie $\widetilde{\vec{h}}_p$ and $\widetilde{\vec{h}}_e$).

We consider to use a fusion network to balance the contributions derived from different types of information, and employ a prediction network to select the response emotion categories based on such mixed information. Specifically, we concatenate the obtained $\widetilde{\vec{h}}_p$ and $\widetilde{\vec{h}}_e$ and feed them into a sigmoid layer to yield a trade-off weight:
\begin{equation}
w = \sigma([\widetilde{\vec{h}}_p;\widetilde{\vec{h}}_e]),
\end{equation}
The final representation is a weighted sum of the non-linear transformed hidden states of two encoders:
\begin{gather}
\widetilde{\vec{h}}_p^{'} = tanh(\widetilde{\vec{h}}_p),\\
\widetilde{\vec{h}}_e^{'} = tanh(\widetilde{\vec{h}}_e),\\
\widetilde{\vec{h}}_{pe} = w \otimes \widetilde{\vec{h}}_p^{'} + (1-w) \otimes \widetilde{\vec{h}}_e^{'},
\end{gather}
where $\otimes$ indicates element-wise multiplication. The final representation is fed into a prediction network to produce an emotion vector for generation, which is passed through MLP and then mapped into a probability distribution over the emotion categories:
\begin{align}
\oldhat{\vec{e}_r} = \sigma(\vec{W_r}\widetilde{\vec{h}}_{pe} + b),\\
\mathcal{L}_{r} = - \vec{e}_r {\rm log}(\oldhat{\vec{e}_r}),
\end{align}
where $\vec{e}_r$ is the multi-hot representation of the response emotion vector. $\oldhat{\vec{e}_r}$ is the final response emotion vector generated through the proposed emotion selector, which is then passed to the generator for emotional response generation.

However, during the first several training epochs, we noticed that the model can not effectively learn to predict the response's emotion. This may due to the fact that a post can be respond in multiple response (\aka the one-to-many mapping problem). Hence, we assume that with the guidance from the target, it is easier for the emotion selector to find the optimal responding emotion and learn the emotion interaction pattern better. We exploit the guidance from the target by using the recognition network, which is composed of the recognition encoder and the intermediate encoder, and\textbf{ we apply the same method of a fusion-prediction network to obtain $\widetilde{\vec{h}}_{re}$ and $\mathcal{L}_{r'}$}.

In training stage, we optimize both the prior and recognition network, but only feed the emotion vector predicted from the recognition network into the response generator. As the target is not available in testing stage, prior network serve the purpose of predicting the emotion vector, and recognition network is no longer used in such time. An intuitive idea of minimizing the discrepancy between training and testing stage is to apply Kullback-Leibler divergence loss between the two predicted emotion vectors:
\begin{equation}\label{Eqn:25}
\mathcal{L}_{KL} = \sum_{i=1}^{K}p(\vec{e_r}=e_i|\vec{x},\vec{y})log \frac{p(\vec{e_r}=e_i|\vec{x},\vec{y})}{p(\vec{e_r}=e_i|\vec{x},\vec{e_p})}.
\end{equation}
However, as the emotion vectors are already restrained by cross-entropy loss, instead of optimizing Eqn. \ref{Eqn:25} here we apply KL-divergence on the hidden states of $\widetilde{\vec{h}}_{pe}$ and $\widetilde{\vec{h}}_{re}$ to reduce the distribution divergence in hidden space:
\begin{equation}
\mathcal{L}_{KL} = KL(\sigma(\vec{W_{kl}}\widetilde{\vec{h}}_{pe} + b)||\sigma(\vec{W_{kl}}\widetilde{\vec{h}}_{re} + b)).
\end{equation}
Note that the intermediate encoder is used in both prior network and recognition network because the asymmetry in their network would widen the gap between them, adding more trouble into optimizing the whole neural network.

Intuitively, the recognition network provides rich guidance from the target to choose the optimal responding emotion and fusion network can adaptively determine the weight between different type of information, ensuring accurate prediction of response’s emotion.

\subsection{Emotion-Biased Response Generator}
To construct the generator, we consider to use an emotion-enhanced seq2seq model that is capable of balancing the emotional part with the semantic part and generate intelligible responses.

Thereby, we first generate the response emotion embedding $\vec{V}_e$ by multiplying $\oldhat{\vec{e}_r}$ with a randomly initialized matrix:
\begin{equation}
\vec{V}_e = \matrix{W}_e\oldhat{\vec{e}_r},
\end{equation}
where $\matrix{W}_e$ is the emotion category embedding matrix, which is a high-level abstraction of emotion expressions. Note that we follow Plutchik's assumptions~\shortcite{plutchik1980general} about the complicated nature of human emotion and believe it is possible that response sentence contains several diverse emotions. Thus here we do not use a softmax on $\oldhat{\vec{e}_r}$ to only pick only ONE optimal emotion category for generation. As such, we call it as \emph{soft-emotion injection} procedure, which is used to model the diversity of response emotions.

By following the work \cite{vinyals2015grammar}, we use a new encoder to encode $\vec{x}$ for obtaining a sequence of hidden states $\vec{h} = (\vec{h}_1,\vec{h}_2,\cdots, \vec{h}_T)$ through a RNN network,
and then generate the context vector $\vec{c}_t$ for decoding the current hidden state $\vec{s_t}$, via applying attention mechanism to re-assign an attended weight to each encoder hidden state $\vec{h}_i$.
\begin{eqnarray}
u^t_i &=& \vec{v^\top} \tanh (\vec{W}_1 \vec{h}_i + \vec{W}_2 \vec{s}_t), \label{eqn:attn1} \\
a^t_i &=& \mathrm{softmax}(u^t_i) \label{eqn:attn2},\\
\vec{c}_t &=& \sum_{i=1}^{T} a^t_i  \vec{h}_i.
\end{eqnarray}

At each time step $t$, the context vector encoded with attention mechanism enable our model to proactively search for salient information which is important for decoding over a long sentence. However, it neglects the emotion ($\vec{V}_e$) derived from the response during generation, and thus we propose an emotion-biased attention mechanism to rewritten Eq.(\ref{eqn:attn1}),
\begin{equation}
u^t_i = \vec{v^\top} \tanh (\vec{W}_1 \vec{h}_i + \vec{W}_2 \vec{s}_t + \vec{W}_3 \vec{V}_e).
\end{equation}
The context vector $\vec{c}_t$ is concatenated with $\vec{s}_t$ and forms a new hidden state $\vec{s}_t^{'}$:
\begin{equation}
\vec{s}_t^{'} = \vec{W}_{4}[\vec{s}_t ; \vec{c}_t],
\end{equation}
from which we make the prediction for each word;
$\vec{s}_t$ is obtained by changing Eq.~\eqref{eq:over_opt} into:
\begin{equation}
\vec{s}_t = \mathbf{GRU}(\vec{s}_{t-1}^{'}, [ \vec{y}_{t-1} ; \vec{V}_e ]) ,
\end{equation}
which fulfills the task of injecting emotion information while generating responses. To be consistent with previous conversation generation approaches, here we consider to use \emph{cross entropy} to be the loss function, which is defined by
\begin{align}
\mathcal{L}_{NLL}(\theta)&= -logP(\vec{y|x}) \notag  \\
&= -\sum_{t=1}^{T^{'}} log P(y_t \mid y_1, y_2,\cdots, y_{t-1}, \vec{c}_t,\vec{V}_e ) ,
\end{align}
where $\theta$ denotes the parameters.

\subsection{Loss Function}

The loss function of our model is a weighted summation of the \emph{semantic loss} and \emph{emotional loss}:
\begin{equation}
\mathcal{L}_{TG-EACM}(\theta) = \alpha \mathcal{L}_e + (1-\alpha)\mathcal{L}_{NLL},
\end{equation}
where $\alpha$ is a balance factor, and $\mathcal{L}_e$ denotes the \emph{emotional} loss, namely,
\begin{equation}
\mathcal{L}_e = \mathcal{L}_p + \mathcal{L}_{r}+ \mathcal{L}_{r'}+ \mathcal{L}_{KL}.
\end{equation}

\section{Experimental Results}
\label{sec:exp}

\subsection{Dataset}

\renewcommand\arraystretch{1.4}
\begin{table}[!t]
\centering\caption{ Statistic of ESTC. }
\begin{tabular}{|c|c|c|c|}
\hline
\multirow{4}{*}{\textbf{Training $\#$}} & \multicolumn{2}{c|}{\textbf{Posts}}                     & 219,162   \\ \cline{2-4}
                          & \multirow{3}{*}{\textbf{Responses}} & \textbf{\textit{No Emotion}} & 1,586,065 \\ \cline{3-4}
                          &                            & \textbf{\textit{Single Emotion}  } & 2,792,339 \\ \cline{3-4}
                          &                            & \textbf{\textit{Dual Emotion} }& 53,545    \\ \hline
\textbf{Validation $\#$}                & \multicolumn{3}{c|}{1,000}                                 \\ \hline
\textbf{Testing $\#$}                   & \multicolumn{3}{c|}{1,000}                                 \\ \hline
\end{tabular}
\label{tab:2}
\end{table}

\begin{figure}
    \begin{minipage}{0.485\linewidth}
      \centerline{\includegraphics[scale=0.25]{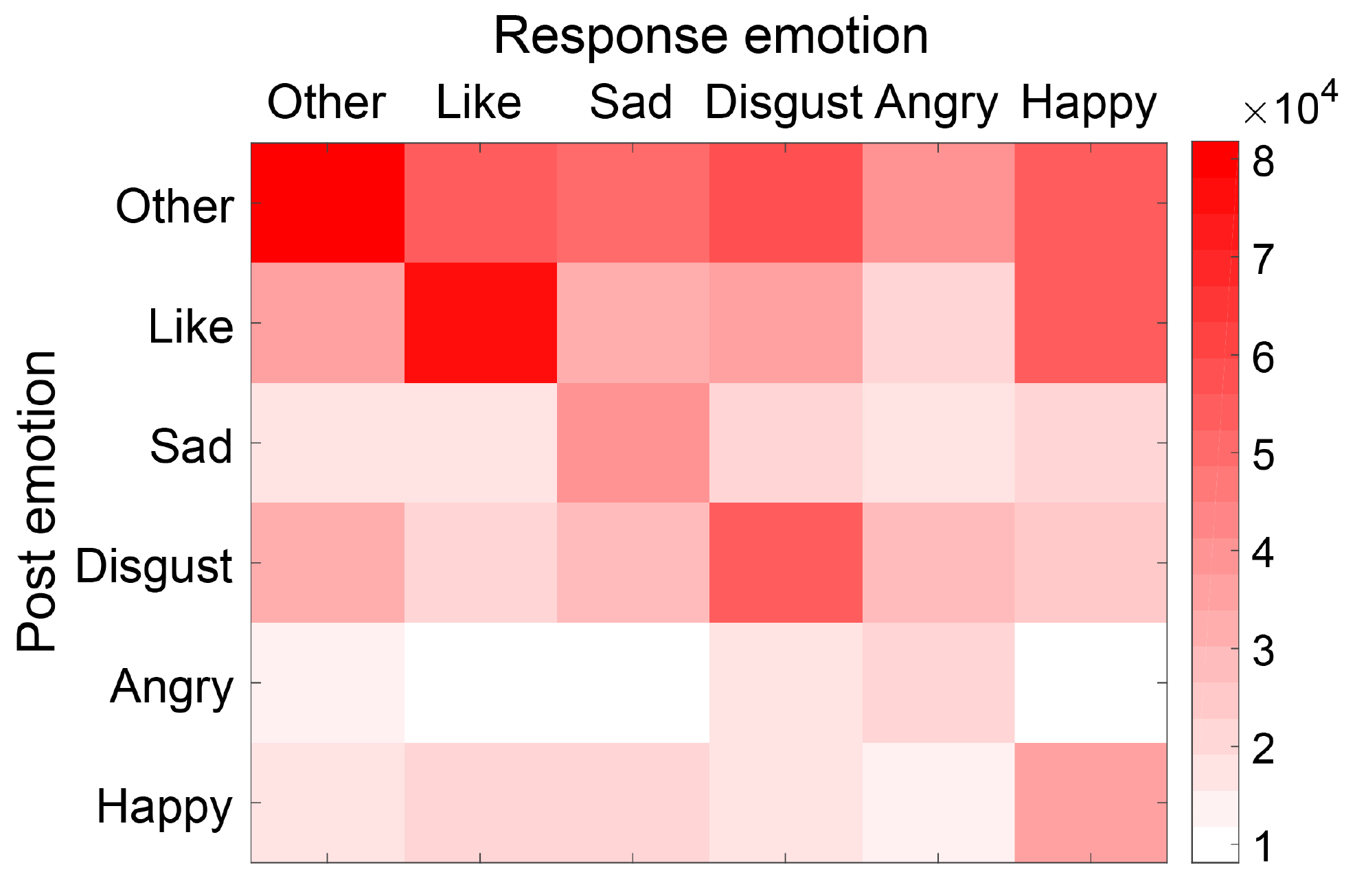}}
      \centerline{(a) EIPs on primary label.}
    \end{minipage}
    \hfill
    \begin{minipage}{0.485\linewidth}
      \centerline{\includegraphics[scale=0.35]{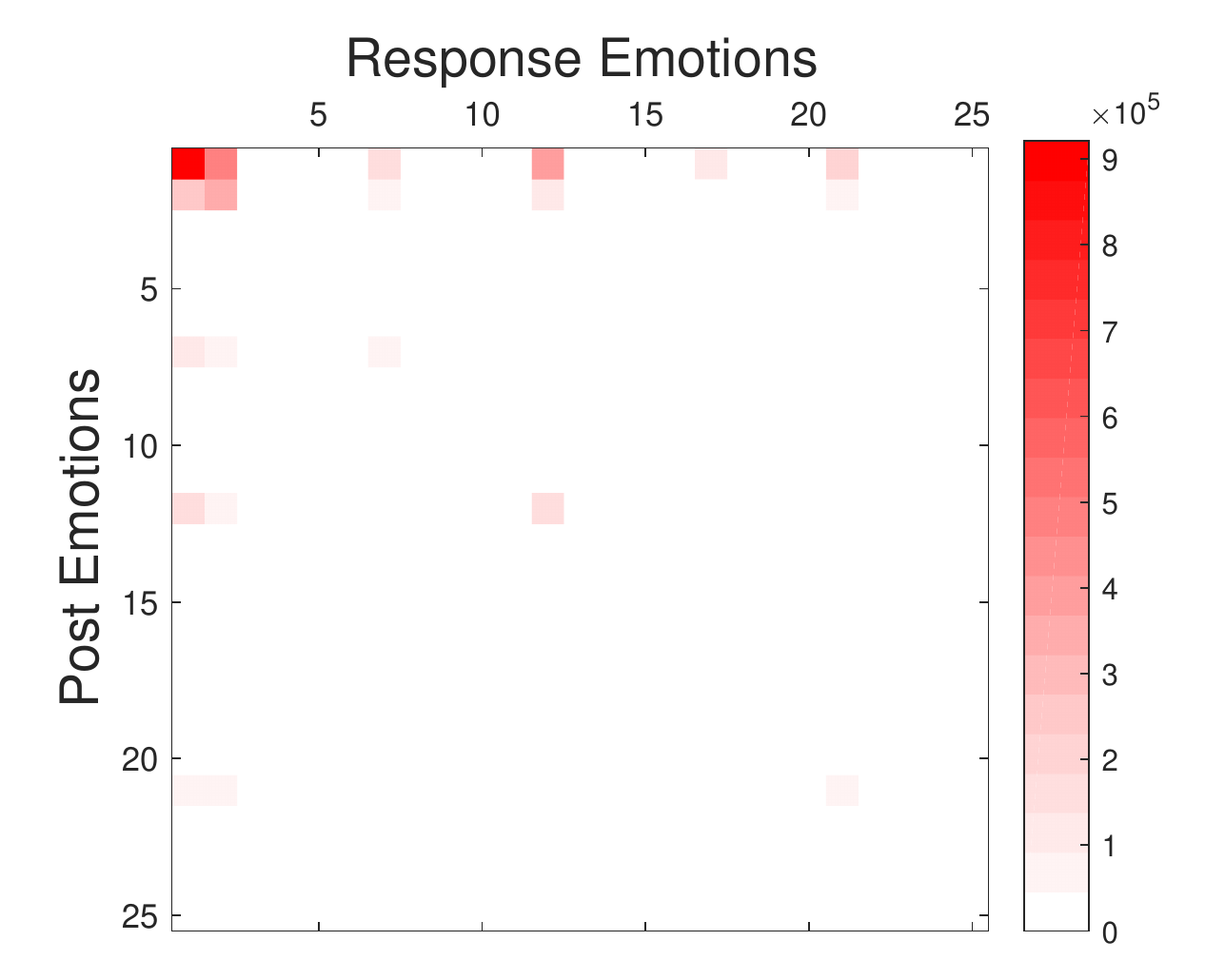}}
      \centerline{(b) EIPs on dual label.}
    \end{minipage}
    \vfill
    \caption{An illustration of emotion interaction pattens (EIPs) considering only primary emotion or both primary and secondary emotion label. The darker each grid is, the more pairs ($\vec{e_p}$,$\vec{e_r}$) appear in the dataset. Need to note that in pic.(b) the labels are replaced by numerical numbers for simplicity.}
    \label{fig4}
\end{figure}

A large-scale publicly available dataset, \ie Emotional Short-Text Conversation (ESTC), is used in the experiments to evaluate the performance of the models. This dataset is  derived from STC \shortcite{shang2015neural} and contains over four million real-world conversations collected from Chinese Weibo, a social media platform. The raw dataset contains $4,433,949$ post-comment pairs, from which $1,000$ pairs are extracted for validation and another $1,000$ pairs are used for testing. Details of ESTC are shown in Table \ref{tab:2}.

\paratitle{Preprocessing}.
As the raw dataset (STC) does not have emotion labels, we follow the work~\cite{zhou2018emotional} to train an emotion classifier for assigning emotional labels to the sentences in the dataset. Specifically, we train an BERT-based \cite{devlin2018bert} emotion classifier  model over two different datasets, \ie emotion classification datasets in NLPCC 2013\footnote{\url{http://tcci.ccf.org.cn/conference/2013/}} and NLPCC 2014\footnote{\url{http://tcci.ccf.org.cn/conference/2014/}}, which contain $29,417$ manually annotated data in total, and each sentence in such datasets is marked with two emotional labels, namely, a primary label and a secondary one, and thus we regard it as a multi-label classification task. Specifically, we preprocess the labels over the mentioned emotion categories, \ie (\emph{like, disgust, sad, angry, happy, other}) and remove the rare ones (\eg fear). Note that here ``\emph{other}'' indicates no emotion. Under such circumstances, the sentences in the dataset can be roughly categorized into three classes, \ie without emotions (other, other), with a single emotion (other, other) and with dual emotions (emo1, emo2). In addition, we also train a BERT model over the pre-processed dataset for input, and the performance (\ie accuracy) of the trained BERT model is $72.57\%$.

To evaluate the emotion perception ability of different approaches over the emotion categories, we build an emotion-rich dialogue \emph{test} set for a fair empirical comparison. Specifically, we randomly choose $1,000$ post-response pairs whose primary emotion label is among the above categories (\ie \emph{like, disgust, sad, angry or happy}), each of which contains $200$ pairs. Additionally, we also present an in-depth analysis on \emph{emotion interaction patterns} (EIPs) over conversations,
in which each EIP is defined as a ($\vec{e_p}$,$\vec{e_r}$) pair for each conversation to indicate the transition pattern from the post's emotion to the response's one.
Figure \ref{fig4} shows the heat-map shows a heatmap to depict the number of EIPs appearing in the dataset, and each row (\emph{or} column) denotes the post's emotion $\vec{e_p}$ (\emph{or} the response's emotion $\vec{e_r}$). From Figure \ref{fig4} we can observe that: (1) The darker each grid is, the more ($\vec{e_p}$,$\vec{e_r}$) pairs appear in the dataset; (2) The heat map in  Fig~\ref{fig4}.(b) is sparse since the EIPs of some post-response emotion pairs, \eg (happy, sad), are overly rare to appear in our dataset, and thus the column (or row) in Fig~\ref{fig4}.(b) is $25$ (not $36$); (3) For ease of analysis, we also provide the heat-map only containing the primary emotion labels of post-response pairs in Fig~\ref{fig4}.(a).

Moreover, we also leverage ``Weibo Sentiment Dataset'' provided by Shujutang\footnote{\url{http://www.datatang.com/data/45439}} to train the sentiment-specifi word embeddings (SSWE), which consists of two million Weibo sentences with sentiment labels, and we remove some extra domain-specific punctuations like ``@user'' and ``URLs''.

\subsection{Evaluation Metric}
As mention in \cite{liu2016not}, the automatic evaluation of generation models still remains challenges. In this paper, we follow the previous work \cite{zhou2018emotional} to adopt  {\it distinct-1}, {\it distinct-2}~\cite{li2015diversity}, {\it BLEU-1} and {\it BLEU-2}~\cite{papineni2002bleu} to automatically evaluate our method and all baselines. In addition, $3$ graduate students (whose research areas are not in text processing area) are invited for manually labeling the generated responses, which are scored according to both \emph{emotional}-level and \emph{semantic}-level information, and then the {\it response quality} is calculated by combining the scores from such two levels for integrally evaluating all methods.

\paratitle{Automatic Evaluation}.
Here we use {\it distinct-1}, {\it distinct-2}~\cite{li2015diversity}, {\it BLEU-1} and {\it BLEU-2}~\cite{papineni2002bleu} to be the automatic evaluation metrics.
Specifically, {\it Distinct-n} measures the degree of diversity by computing the number of distinct n-grams in the generated responses and can indirectly reflect the degree of emotion diversity, as the generated sentence containing diverse emotions is more likely to have more abundant words in principle. {\it BLEU-n} is a referenced metric which calculates the n-gram word overlap between the generated response and the target. It reflects how well the model can learn to respond from the training set.

\paratitle{Human Evaluation}.
We randomly sampled $200$ posts from the \emph{test} set, and then aggregate the corresponding responses returned by each to-be-evaluated method, then three graduate students~(whose research focus is not in text processing area) are invited for labeling. Each generated response is labeled from two different aspects, \ie \emph{emotional aspect} and \emph{semantic aspect}. Specifically, from emotional perspective, each generated response is labeled with $0$ if its emotion is apparently inappropriate (namely evident emotion collision, \eg angry-happy), and $1$ for otherwise. From semantic perspective, we evaluate the generated results using the scoring metrics as follows. Note that if conflicts happen, the third annotator determines the final result.

\begin{itemize}
    \item 1: If the generated sentence can be obversely considered as a appropriate response to the input post;
    \item 0: If the generated sentence is hard-to-perceive or has little relevance to the given post.
\end{itemize}

To conduct an integral assessment of the models at both \emph{emotional}-level and \emph{semantic}-level, we measure the {\it response quality} by using the formula as follows,
\begin{equation}
Q_{response} = S_{emotion} \land S_{semantics},
\end{equation}
where $Q_{response}$ reflects the {\it response quality} and $S_{emotion}$, $S_{semantics}$ denote the emotional score and semantic score, respectively. The {\it response quality} of each case is equal to 1 if and only if both of its emotion score and semantic score are scored as 1.

\subsection{Baselines}

We compare our model {(TG-EACM)} with the following baselines.

\paratitle{Seq2seq}~\cite{sutskever2014sequence,vinyals2015grammar}. The canonical {\it seq2seq} model with attention mechanisim.

\paratitle{ECM}~\cite{zhou2018emotional}. {\it ECM} is improper to directly be as a baseline if without a pre-assigned emotion category.

Thereby, we manually designate a most frequent response emotion to {\it ECM} for fair comparison. Specifically, we train a post emotion classifier to automatically detect post's emotion, and then choose the corresponding response emotion category using the most frequent response's emotion to the detected post's emotion over EIPs, which is easy to find according to Figure \ref{fig3}.

\paratitle{Seq2seq-emb}~\cite{zhou2018emotional,huang2018automatic}. \emph{Seq2seq} with emotion embedding ({\it Seq2seq-emb}) is also adopted in the same manner. This model encodes the emotion category into an embedding vector, and then directly utilizes it as an extra input when decoding (\ie \emph{hard emotion injection}).

Intuitively, the generated responses from {\it ECM} and {\it Seq2seq-emb} can be regarded as the indication of the performance of simply incorporating the \emph{EIP}s,
to model the emotional interactions over the conversation pairs. For evaluating the impact of recognition network, we directly use the target responses when training,
which is used for investigate the effectiveness of the guidance information derived from the ground truth, and how it can facilitate the model
during the emotion-aware response selection-generation process, and we also add a variant of our proposed model for comparison, which iss constructed by removing the recognition network from \ie {\it TG-EACM} (\ie only using the prior network for test).

\subsection{Implementation Details}
For all approaches, each encoder and decoder with $2$-layers GRU cells containing 256 hidden units, and all of the parameters are not shared between such two different layers. The vocabulary size is set as $40,000$, and the OOV (out-of-vocabulary) words are replaced with a special token {\it UNK}. The size of word embeddings is $200$, which are randomly initialized. The emotion category embedding is a $6\times 200$-dimensional matrix (if used). The parameters of {\it imemory} and {\it ememory} in {\it ECM} are the same as the settings in  \cite{zhou2018emotional}. We use stochastic gradient descent (SGD) with mini-batch for optimization when training, and the batch size and the learning rate are set as $128$ and $0.5$, respectively. The greedy search algorithm is adopted for each approach to generate responses. Additionally, for speeding up the training process, we leverage the well-trained {\it Seq2seq} model to initialize other methods.

For our proposed method, the parameters are empirically set as follows: \emph{SSWE} is trained by following the parameter settings in~\cite{tang2014learning}, where the length of hidden layer is set at $20$ and window size at $3$, and AdaGrad \cite{Duchi2011Adaptive} is used to update the trainable parameters with the learning rate of $0.1$. The size of emotional embedding and semantic embedding are both set at $200$. In particular, the \emph{Word2vec} embedding is used based on {\it Tencent AI Lab Embedding\footnote{\url{https://ai.tencent.com/ailab/nlp/embedding.html}}}, which is pre-trained over $8$ million high-quality Chinese words and phrases by using directional skip-gram method \cite{song2018directional}. We use {\tt jieba\footnote{\url{https://github.com/fxsjy/jieba}}} for word segmentation during the evaluation process.

\subsection{Results and Discussion}
\label{subsec:results}
In this section, we evaluate the effectiveness of generating emotional responses by our approach as comparison to the baseline methods.

\paratitle{Automatic Evaluation}.
From Table \ref{tab3}, we can observe that: (i) \emph{ECM} performs worse than \emph{Seq2seq}, the reason might be that the generation process is a two-stage process, \ie post emotion detection process and response generation process, which would significantly reduce the diversity and quality of emotion response generation due to the errors of emotion classification and the one-to-one mapping of emotional interaction pattern. In particular, the emotion category {\it (other, other)} is more likely to be chosen than other emotion categories. We will present an in-depth analysis in Section~\ref{sec:exp-case-study}. (ii) Our proposed \emph{TG-EACM} consistently outperforms all of the baselines in terms of all the metrics and the improvements are statistically significant (p-value $<0.05$). The results demonstrate that our target-guided emotion selection process is really effective in enhancing the ability of generating more diverse words and high-quality responses. Deep and thorough analysis between \emph{TG-EACM} and \emph{EACM} will be presented in Section \ref{sec:ablation}.

\begin{table}[!htp]
\centering\caption{Automatic evaluation: Distinct-n and BLEU-n}
\begin{tabular}{lcccc}
\hline
\textbf{Models}      & \textbf{Distinct-1} & \textbf{Distinct-2} & \textbf{BLEU-1} & \textbf{BLEU-2} \\
\hline
\textbf{Seq2seq}     & 0.0666              & 0.2147              & 0.2429          & 0.1589         \\
\textbf{Seq2seq-emb} & 0.0691              & 0.2426              & 0.2229          & 0.1470            \\
\textbf{ECM}         & 0.0603              & 0.2070              & 0.2316          & 0.1525           \\
\textbf{EACM}        & 0.0819              & 0.2840              & 0.2305          & 0.1518        \\
\textbf{TG-EACM}     & \textbf{0.0839}     & \textbf{0.4070}     & \textbf{0.2708} & \textbf{0.1832}   \\
\hline
\end{tabular}
\label{tab3}
\end{table}

\renewcommand\arraystretch{1.2}
\begin{table}[!htp]
\centering\caption{Human evaluation: averaged semantic score, emotional score and response quality.}
\begin{tabular}{lccc}
\hline
\textbf{Models}      &\textbf{$\bar{\bm{S}}_{\text{semantics}}$} &\textbf{$\bar{\bm{S}}_{\text{emotion}}$}
 & \textbf{$\bar{\bm{Q}}_{\text{response}}$}  \\ \hline
\textbf{Seq2seq}     & 0.390          & 0.815           & 0.360        \\
\textbf{Seq2seq-emb} &    0.280       &  0.795         & 0.250         \\
\textbf{ECM}         & 0.355          & 0.870  & 0.310         \\
\textbf{EACM}         & \textbf{0.415}          & 0.885  & \textbf{0.390}          \\
\textbf{TG-EACM}        & 0.410 & \textbf{0.910}  & \textbf{0.390}      \\ \hline
\end{tabular}
\label{tab4}
\end{table}

\paratitle{Human Evaluation}.
Results for human evaluation are listed in Table \ref{tab4}. Here, we calculate inter-rater agreement among the three annotators with the Fleiss\' kappa \cite{fleiss1973equivalence}. The Fleiss\' kappa for semantics and emotion are $0.492$ and $0.823$, indicating ``Moderate agreement'' and ``Substantial agreement'', respectively. Besides, from Table \ref{tab4} we can observe that: 
\emph{Seq2seq-emb} provides the worst performance as expected, this is because the generation process's apparently interrupted by injecting the emotion category embedding, which would significantly reduce the quality of generated responses and thus generate some hard-to-perceive sentences;
\emph{ECM} achieves a relatively better result, as it models the dynamic emotional expressing process using (\ie \emph{internal memory}), which alleviates the problem brought by \emph{hard emotion injection}. In addition, we can see an apparent increase of emotion score for \emph{ECM} model over \emph{Seq2seq} and \emph{Seq2seq-emb}, because the \emph{ECM} model guarantees the emotional accuracy of generated response by further assigning different generation probabilities to emotional/generic words using \emph{external memory};
Our proposed model \emph{TG-EACM} sees improvements over the baseline methods at both semantic level and emotional level $(\mbox{p-value}\leq0.05)$. For example, \emph{TG-EACM} outperforms \emph{seq2seq} model by $5.12\%$, $11.7\%$ and $8.3\%$ in terms of  \emph{semantic score}, \emph{emotion score} and \emph{response quality}, respectively. Also, as the \emph{response quality} result shows our model is able to balance the emotion selecting process and generating process within a unified model, and thus generating plausible responses with appropriately expressed emotion.
The reason might be the fact that \emph{TG-EACM} is capable of simultaneously encoding the semantics and the emotions in a post and is good at gaining expressing skills from the guidance of the target. Note that comparing to the \emph{EACM}, there's a minor increase in emotion score, which also proves that the target helps the model to select correct emotion. We will do an ablation study to prove the efficacy of our method in the next section.

\begin{table}[!htp]
\centering\caption{The percentage of the emotional-semantic score given by human evaluation.}
\small
\begin{tabular}{lcccc}
\hline
\textbf{Method(\%)}  & 1-1           & 1-0             & 0-1              & 0-0               \\ \hline
\textbf{Seq2seq}     & 36.0          & 45.5          & 3.0            & 15.5          \\
\textbf{Seq2seq-emb} & 25.0          &   54.5        &  3.0           & 17.5           \\
\textbf{ECM}         & 31.0          & \textbf{56.0} & 4.5            & \textbf{8.5}    \\
\textbf{EACM}        & \textbf{39.0} & 49.5          & 2.5            & 9.0               \\
\textbf{TG-EACM}     & \textbf{39.0} & 52.0          &  \textbf{2.0}  & 7.0             \\ \hline
\end{tabular}
\label{tab5}
\end{table}

The percentage of the emotional-semantics scores under the human evaluation is shown in Table \ref{tab5}.
For \emph{ECM}, the percentage of (0-0) declines while the percentage of (1-0) increases as opposed to \emph{Seq2seq}, which suggests that simply using EIP, \ie the most frequent response emotions, has lower probability in causing emotional conflicts and improves emotional satisfaction to a certain extent. However, this amelioration comes at a price, as the percentage of (1-1) decreases because of the fact that directly using EIPs to model emotion interaction process is insufficient for complex human emotions.
In comparison, \emph{TG-EACM} reduces the percentage of generating responses with wrong emotion and correct semantics (\ie 0-1)
while increase the percentage of (1-1) correspondingly, which demonstrates that \emph{TG-EACM} is capable of successfully modeling the emotion interaction pattern among human conversations and meanwhile can guarantee the semantic correctness in sentences. Moreover, it can be concluded that the information from the target help enhance emotion selecting ability, as the percentage of (1-0) increases when comparing \emph{TG-EACM} with \emph{EACM} model.

\subsection{Ablation Study}\label{sec:ablation}
In this section, we conduct an ablation study on evaluating the effectiveness of different components of the proposed model.

\subsubsection{Effectiveness of the guidance information}
As mentioned, the accuracy of emotion prediction might affect the performance of generating appropriate responses. To verify the assumption, we first evaluate the emotion prediction accuracy for the performance of TG-EACM and EACM, which is useful to evaluate the effectiveness of the learnt guidance information derived from the target. Then, we compare the overall performance of our paper and baselines based on the automatic metrics and the human evaluations.

\begin{figure}
\centering
\includegraphics[scale=0.45]{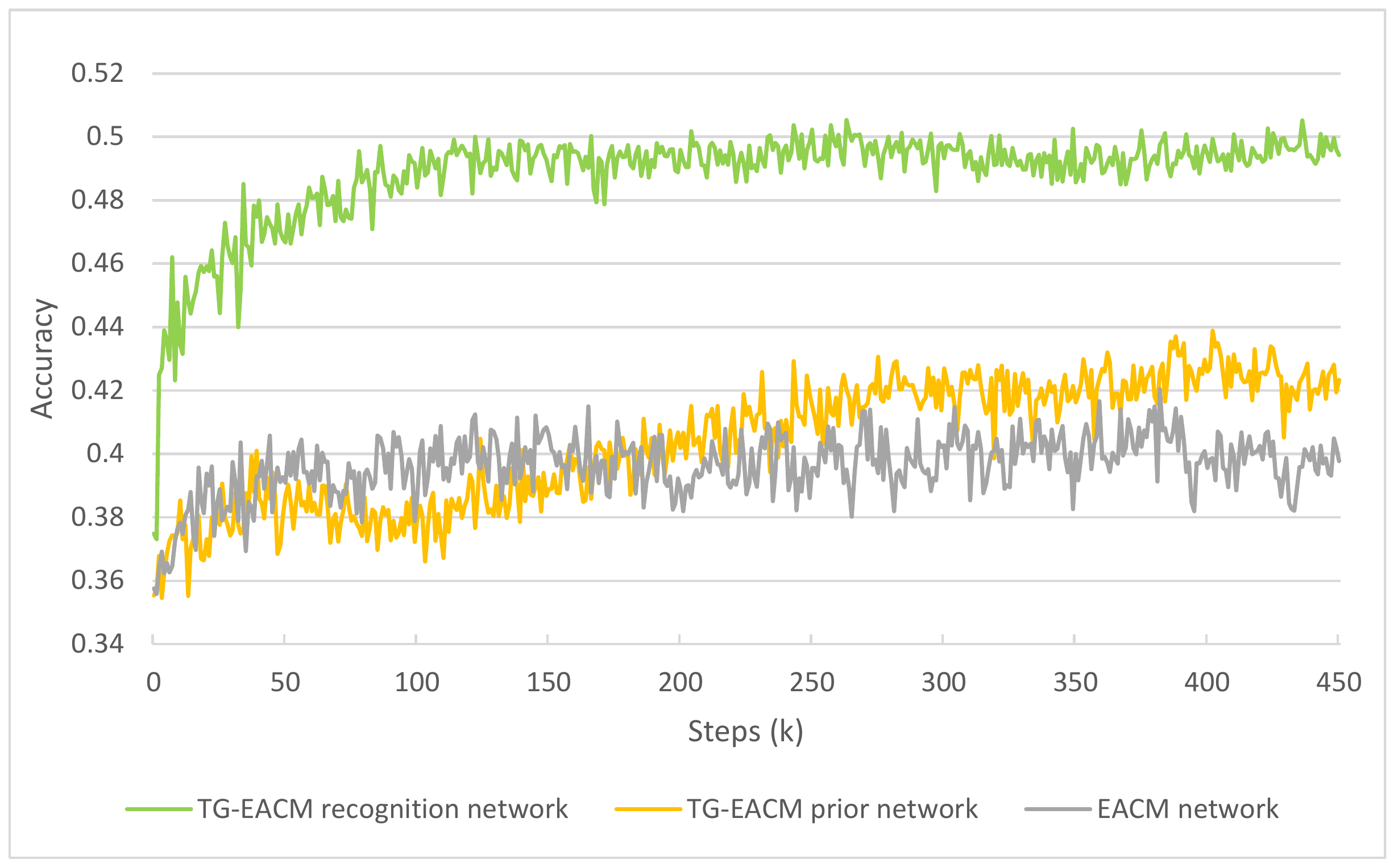}
\caption{The emotion prediction accuracy of TG-EACM and EACM. The X-axis denotes the number of training steps per thousand (k). The prior network of TG-EACM is able to learn  the guidance information of the target via the recognition network.}
\label{fig6}
\end{figure}

The prediction result is presented in Fig \ref{fig6}. The X-axis represents the training steps (per thousand steps, k for brevity), and the Y-axis represents the emotion prediction accuracy. In particular, we will evaluate the prior network and the recognition network of TG-EACM, respectively.
From Fig.~\ref{fig6} and Table \ref{tab6} we can observe that:
(1) At around $100$k steps, the recognition network of TG-EACM  gradually converges and the prior network reaches at its local lowest point;
(2) After $100$k steps, the recognition network fluctuates around $0.49$ ACC, and then the prior network learns from the recognition network and finally reaches its peak value at $0.439$ ACC;
(3) The prediction accuracy of EACM  is higher than the one of the prior network of TG-EACM  at the beginning of training,
and finally the prior network of TG-EACM outperforms EACM. In addition, the peak value of EACM is $0.431$ ACC, and the prior network of TG-EACM performs better than EACM by about $4.38$\%.

\begin{table}[!htp]
\centering\caption{Automatic evaluation on EACM and TG-EACM.}
\footnotesize
\begin{tabular}{lcccc}
\hline
\textbf{Models}      & \textbf{Distinct-1} & \textbf{Distinct-2} & \textbf{BLEU-1} & \textbf{BLEU-2} \\
\hline
\textbf{EACM}        & 0.0819              & 0.2840              & 0.2305          & 0.1518        \\
\textbf{TG-EACM}     & 0.0839              & 0.4070              & 0.2708          & 0.1832     \\
\hline
\textbf{Increment}   & +2.44\%             & +43.31\%            & +17.49\%        & +20.67\%   \\
\hline
\end{tabular}
\label{tab6}
\end{table}

The automatic evaluation and human evaluation results are already shown in Table \ref{tab3} and Table \ref{tab4}. From the results we can observe that the guide information derived from the target is useful for better learning the post-response emotion transition pattern and resulting in generating more interesting and diverse responses.

\begin{figure}
    \begin{minipage}{0.485\linewidth}
      \centerline{\includegraphics[scale=0.32]{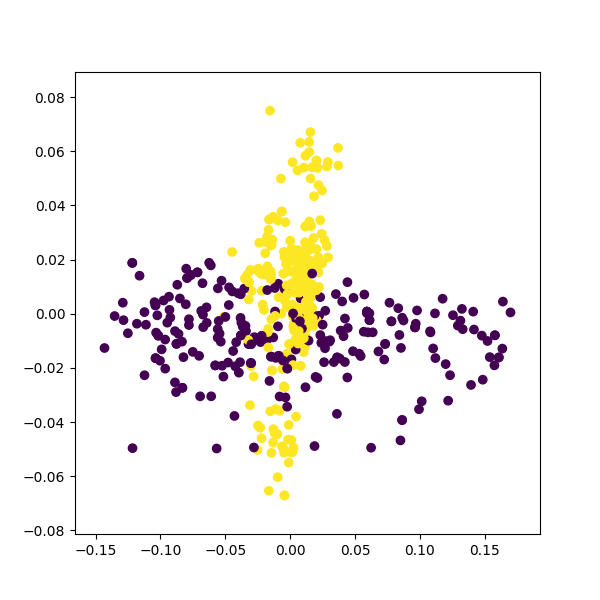}}
      \centerline{(a) Step=1,000}
    \end{minipage}
    \hfill
    \begin{minipage}{0.485\linewidth}
      \centerline{\includegraphics[scale=0.32]{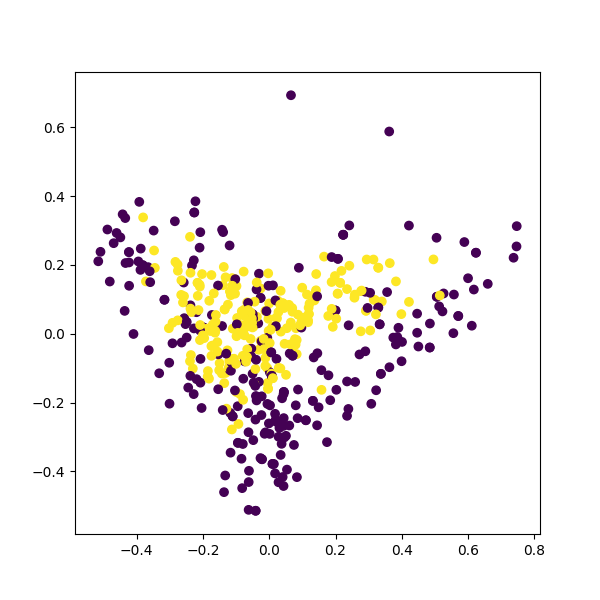}}
      \centerline{(b) Step=420,000}
    \end{minipage}
    \vfill
    \caption{The sampled response emotional distribution in the validation set at different training steps. Yellow dots correspond to samples from prior distribution, while the purple dots correspond to samples from posterior distribution.}\label{fig5}
\end{figure}

In addition, we present  an in-depth analysis of the emotion distribution of the results generated by the prior network and the recognition network of our proposed model, by sampling  $500$ instances from their  the validation sets during training. For ease of analysis, we utilize Principal Components Analysis (PCA) algorithm to handle the emotional distributions for dimensionality reduction, and the results at different training steps are summarized in Figure \ref{fig5}, from which we can observe that with the constraints of KL-loss and emotional Cross-Entropy loss, the prior emotion distribution is gradually close to the posterior emotion distribution along with the training process, which demonstrates the effectiveness of our proposed prior-posterior learning framework.

\subsubsection{Effectiveness of pre-training and word embeddings}
\begin{figure}[!t]
\centering
\includegraphics[scale=0.35]{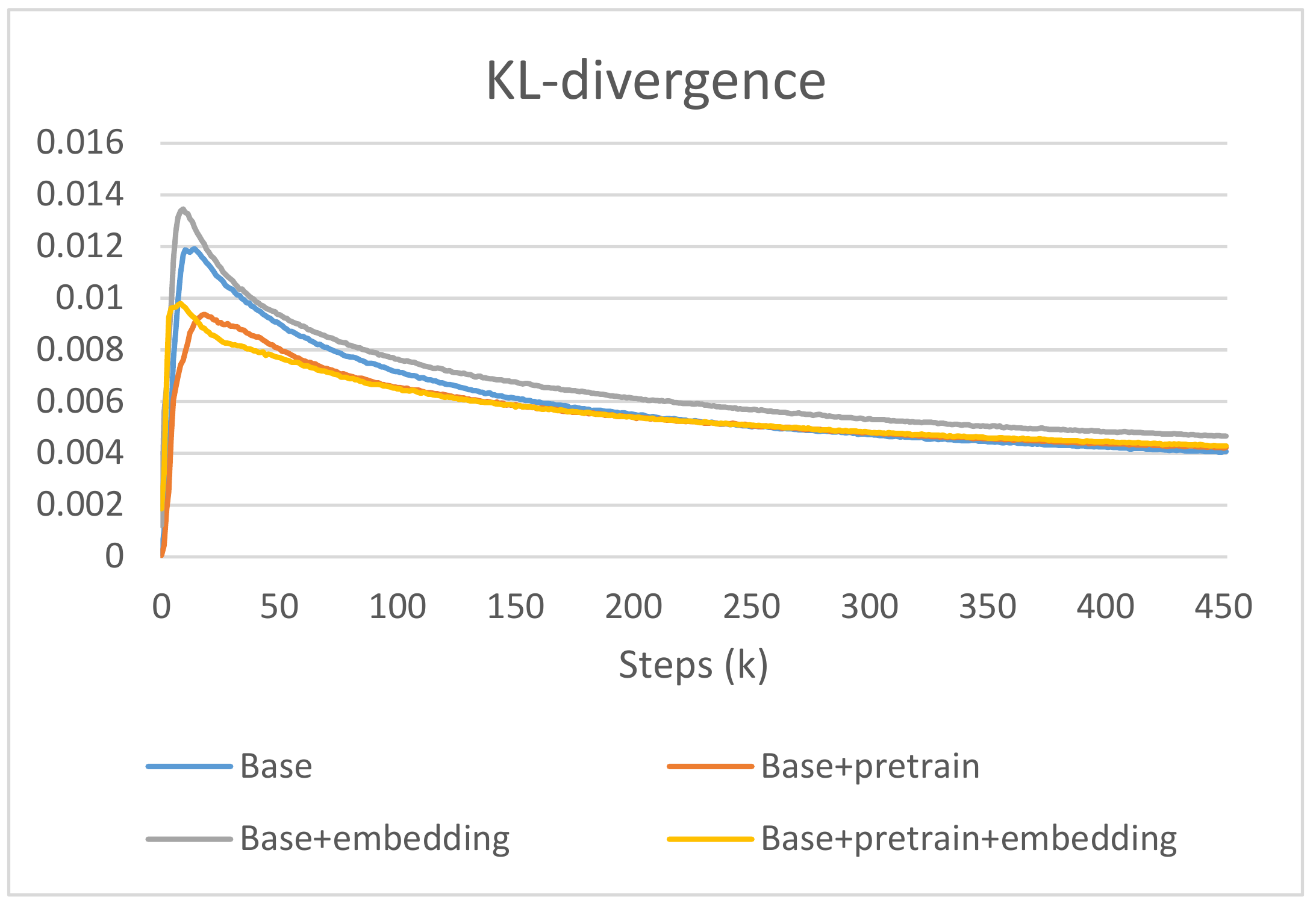}
\caption{The value of KL-divergence term in training period. The X-axis denotes number of thousand training steps.}
\label{fig7}
\end{figure}

\begin{figure}
    \begin{minipage}{0.485\linewidth}
      \centerline{\includegraphics[scale=0.28]{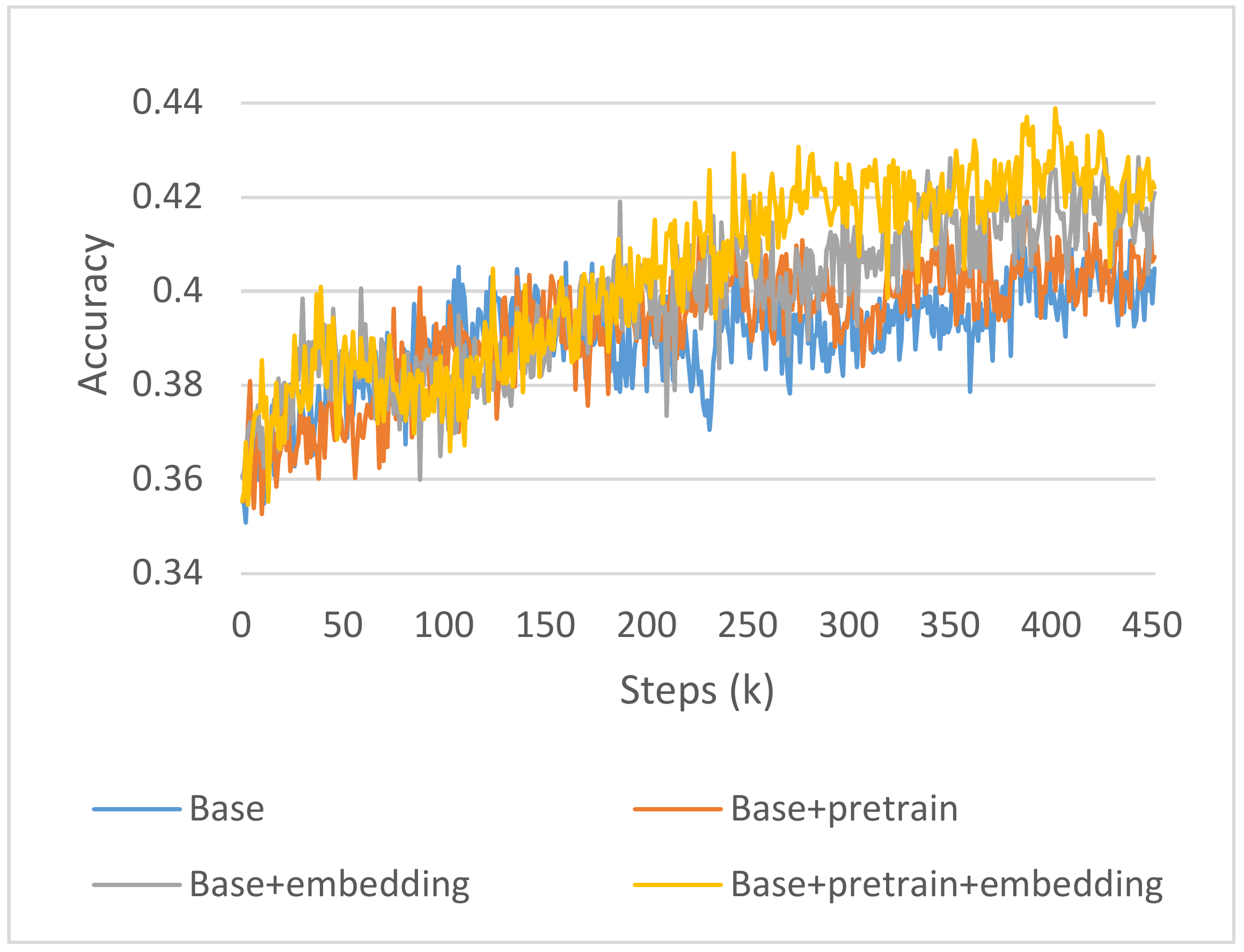}}
      \centerline{(a) Original curve}
    \end{minipage}
    \hfill
    \begin{minipage}{0.485\linewidth}
      \centerline{\includegraphics[scale=0.29]{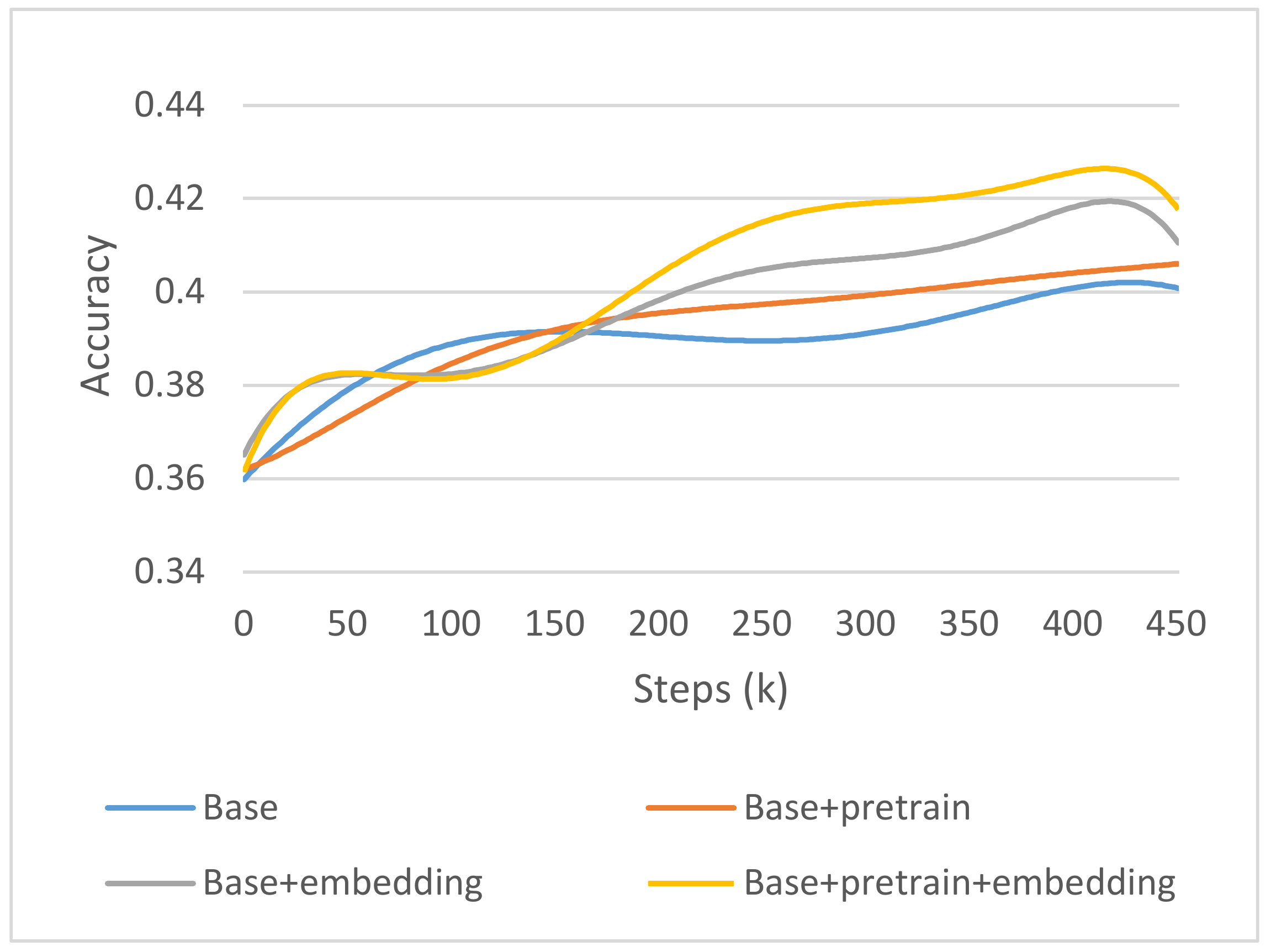}}
      \centerline{(b) Polynomial fitting curve}
    \end{minipage}
    \vfill
    \caption{The emotional prediction accuracy training curve from prior network. The X-axis denotes number of thousand training steps.}\label{fig8}
\end{figure}

In this subsection, we evaluate the impact of the diverse word embeddings (\ie SSWE embedding and Word2Vec embedding), as well as the effectiveness of pre-training. To evaluate the performance of our proposed model in terms of KL-divergence and emotional accuracy during training, the results of which is given in Fig. \ref{fig7} and \ref{fig8}, where the basic model is generated according to \emph{TG-EACM} without using \emph{seq2seq} for pre-training, and applied randomly-initialized word embeddings for  the encoders and decoders. From Fig. \ref{fig7} we can observe that: 
(1) SSWE embedding and Word2Vec embedding could  lead to convergence quickly;
(2) Using the trained \emph{seq2seq} model to initialize the parameters of our model can significantly reduce the loss of KL-divergence.
To this end, the above two observations can be jointly used for speeding up the training process. In addition, from Fig. \ref{fig8}, we can approximate the original emotion accuracy curve through six-time polynomial fitting, which also indicates the efficacy of the above two tricks.

\begin{figure*}[!htp]
\centering
\includegraphics[scale=0.53]{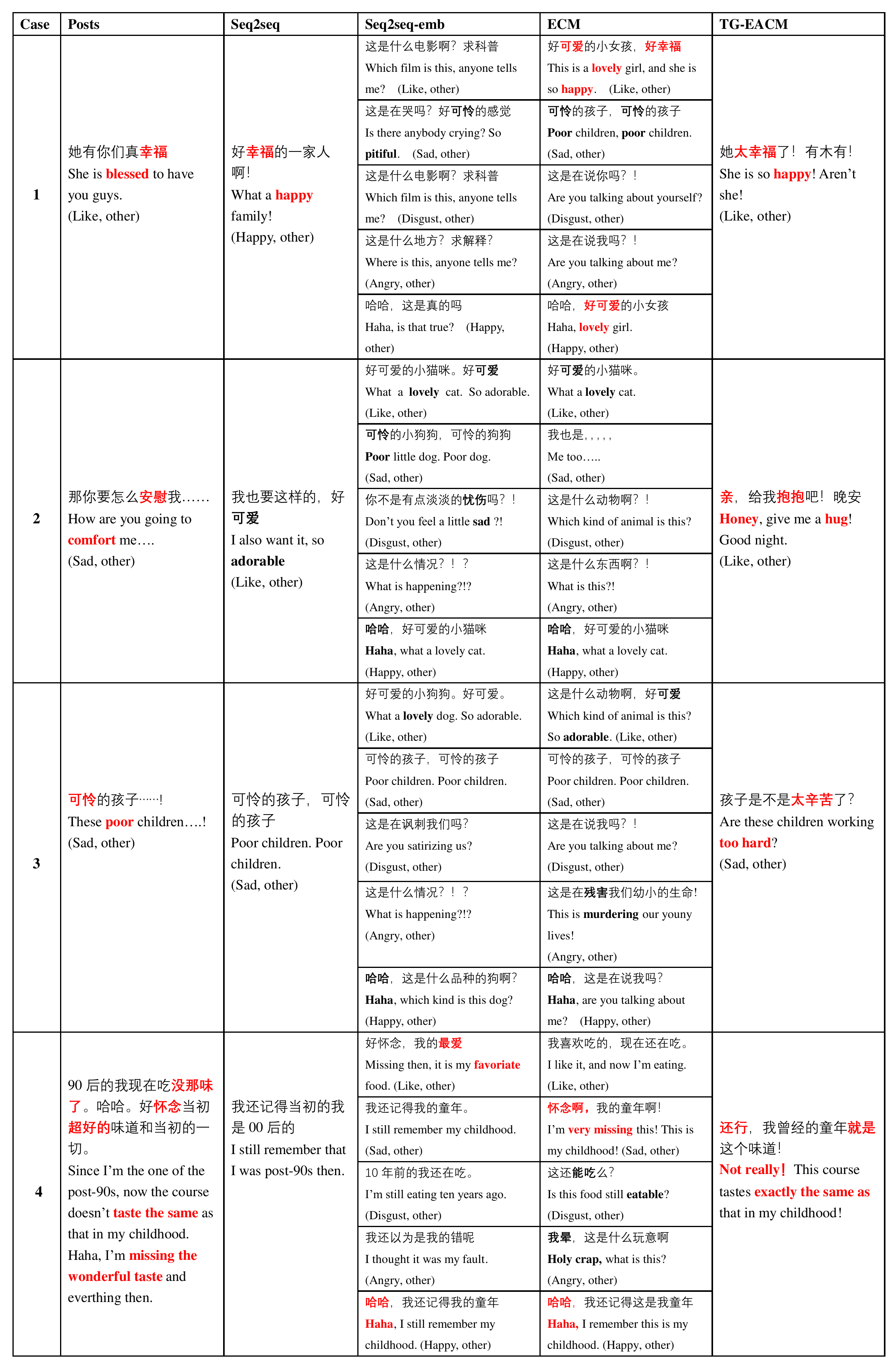}
\caption{Case Study. Three Samples (with the given posts and the corresponding responses) generated by \emph{Seq2seq}, \emph{Seq2seq-emb}, \emph{ECM} and \emph{TG-EACM}.
Words that express appropriate emotion in responses are highlighted in red,  along with their posts' corresponding emotion words, while those expressing inappropriate emotion are bold in black.}
\label{fig9}
\end{figure*}

\subsection{Case Study}
\label{sec:exp-case-study}
In this section, we present an in-depth analysis of emotion-aware response generation results of our proposed approach. Three samples (with the input posts and their corresponding responses) are chosen for comparison with the results generated by different models.

\paratitle{Case 1}.
The response generated by \emph{TG-EACM} and \emph{Seq2seq} are both emotionally and semantically correct. For \emph{ECM}, we can observe that not all of generated responses are both correct at emotion-level and semantic-level, except some, \eg  (Like, other) and (Happy, other) emotion, which indicates the performance of \emph{ECM} is affected by the manually selected emotion in practice. Besides, the responses with (Sad, other) emotion generated by \emph{ECM} might be improper for the given post and may cause conflict between the interlocutors. \emph{Seq2seq-emb} fails to caputure the semantic meaning of the post and thus generates the irrelevant response, which demonstrates the efficacy of our guided-based emotion-aware response generation model for automatically generating optimal emotional response.

\paratitle{Case 2}.
In this case, the post is inclined to express a sad feeling and might seek for comfort. However, we can observe that only \emph{TG-EACM} is capable of recognizing and selecting the correct emotion, and thus generate a response with emotional resonance. In addition, \emph{Seq2seq} cannot detect the emotion and thus generates an irrelevant response as expected. All of results generated by ECM with different emotions seems improper for the post (especially for \emph{ECM} with (Happy,other)),
which reflects directly using a single and designated emotion for generation might be insufficient to model the emotion interaction pattern. Moreover, simply using EIPs would fail to capture the post-response emotion transition, such as (Happy, other) to (Sad, other).

\paratitle{Case 3}. The emotion in the case is (Sad, other), however the responses provided by \emph{ECM} and \emph{Seq2seq-emb} with (Sad, other) are just repeating the post. Note that the responses generated by \emph{ECM}  with (Angry, other) is semantically correct, but the response provided by \emph{TG-EACM} is much better due to its more empathetic attitude.

\section{Conclusion}
\label{sec:con}
In this paper, we propose an target guided emotion-aware chat machine (TG-EACM) to address the emotional response generation problem, which is composed of an emotion selector and a response generator. Specifically, the prior/recognition network is used to combine the emotional and semantic information from the post and the guidance information from the target, in order to predict emotion vector which is used to supervise the emotional response generation process. Extensive experiments conducted on a public dataset demonstrate the effectiveness of our proposed method as compared to baselines at both \emph{semantic}-level and \emph{emotional}-level, in terms of automatic evaluation and human evaluation metrics. For future work, we plan to investigate how the classification errors
affect the performance of our proposed model, and study how to model the emotional and semantics distribution of the given post for generating more intelligent responses with appropriately expressed emotions.

\section{Acknowledgement}
This work was supported in part by the National Natural Science Foundation
of China under Grant No. 61602197 and Grant No. 61772076, Grant No. 61972448,
Grant No. L1924068 and in part by Equipment Pre-Research Fund for The 13th Five-Year Plan under Grant No. 41412050801.

\bibliographystyle{ACM-Reference-Format}
\balance
\bibliography{TOIS-acmsmall}

\end{document}